\documentclass[iicol]{sn-jnl}

\usepackage{times}
\usepackage{epsfig}
\usepackage{graphicx}%
\usepackage{multirow}%
\usepackage{bbm}
\usepackage{makecell}
\usepackage{amsmath,amssymb,amsfonts}%
\usepackage{amsthm}%
\usepackage{mathrsfs}%
\usepackage{ulem}
\usepackage[title]{appendix}%
\usepackage{textcomp}%
\usepackage{manyfoot}%
\usepackage{booktabs}%
\usepackage{algorithm}%
\usepackage{algorithmicx}%
\usepackage{algpseudocode}%
\usepackage{listings}%
\usepackage{tablefootnote}
\usepackage{threeparttable}
\usepackage{enumitem}
\usepackage{stfloats}
\usepackage{bm}
\usepackage{bbding}
\usepackage{subfig}
\usepackage{float}
\usepackage[accsupp]{axessibility}
\usepackage[table]{xcolor}
\usepackage[bottom]{footmisc}
\usepackage{fix-cm}

\definecolor{cGreen}{RGB}{100,180,100}
\definecolor{cRed}{RGB}{220,50,0}
\definecolor{Klein_Blue}{rgb}{0.0, 0.129, 0.6}
\definecolor{mygray2}{gray}{0.9}
\definecolor{mygray1}{gray}{0.95}
\definecolor{brown10}{rgb}{0.9, 0.84, 0.80}  
\definecolor{brown20}{rgb}{0.95, 0.90, 0.85}
\newcommand{\ignore}[1]{}
\hypersetup{
    colorlinks=true,
    linkcolor=magenta,
    urlcolor=magenta,
    citecolor=Klein_Blue
    }
\usepackage{CJKutf8}

\begin{document}

\title[Article Title]{Exploiting Lightweight Hierarchical ViT and Dynamic Framework for Efficient Visual Tracking}

\author[1]{\fnm{Ben} \sur{Kang}}\email{kangben@mail.dlut.edu.cn}
\equalcont{These authors contributed equally to this work.}

\author[1]{\fnm{Xin} \sur{Chen}}\email{chenxin3131@mail.dlut.edu.cn}
\equalcont{These authors contributed equally to this work.}

\author*[1]{\fnm{Jie} \sur{Zhao}}\email{zj982853200@dlut.edu.cn}

\author[2]{\fnm{Chunjuan} \sur{Bo}}\email{bcj@dlnu.edu.cn}

\author[1]{\fnm{Dong} \sur{Wang}}\email{wdice@dlut.edu.cn}

\author[1]{\fnm{Huchuan} \sur{Lu}}\email{lhchuan@dlut.edu.cn}

\affil[1]{\orgname{Dalian University of Technology}, \orgaddress{\city{Dalian}, \postcode{116024}, \state{Liaoning Province}, \country{China}}}

\affil[2]{\orgname{Dalian Minzu University}, \orgaddress{\city{Dalian}, \postcode{116024}, \state{Liaoning Province}, \country{China}}}

\abstract{Transformer-based visual trackers have demonstrated significant advancements due to their powerful modeling capabilities. However, their practicality is limited on resource-constrained devices because of their slow processing speeds. To address this challenge, we present HiT, a novel family of efficient tracking models that achieve high performance while maintaining fast operation across various devices. The core innovation of HiT lies in its Bridge Module, which connects lightweight transformers to the tracking framework, enhancing feature representation quality. Additionally, we introduce a dual-image position encoding approach to effectively encode spatial information. HiT achieves an impressive speed of 61 frames per second (fps) on the NVIDIA Jetson AGX platform, alongside a competitive AUC of 64.6\% on the LaSOT benchmark, outperforming all previous efficient trackers.
Building on HiT, we propose DyHiT, an efficient dynamic tracker that flexibly adapts to scene complexity by selecting routes with varying computational requirements. DyHiT uses search area features extracted by the backbone network and inputs them into an efficient dynamic router to classify tracking scenarios. Based on the classification, DyHiT applies a divide-and-conquer strategy, selecting appropriate routes to achieve a superior trade-off between accuracy and speed. The fastest version of DyHiT achieves 111 fps on NVIDIA Jetson AGX while maintaining an AUC of 62.4\% on LaSOT.
Furthermore,  we introduce a training-free acceleration method based on the dynamic routing architecture of DyHiT. This method significantly improves the execution speed of various high-performance trackers without sacrificing accuracy. For instance, our acceleration method enables the state-of-the-art tracker SeqTrack-B256 to achieve a $2.68\times$ speedup on an NVIDIA GeForce RTX 2080 Ti GPU while maintaining the same AUC of 69.9\% on the LaSOT. Codes, models, and results are available at \url{https://github.com/kangben258/HiT}.}


\maketitle
\section{Introduction}\label{sec1}
Visual object tracking is a fundamental task in computer vision that involves continuously tracking a specific object within a video sequence, starting from its initial state. This task has wide-ranging applications in areas such as video surveillance, autonomous driving, and robotic vision. In recent years,  advancements in deep neural networks~\cite{AlexNet,ResNet,2017Attention} have significantly accelerated progress in visual tracking~\cite{li2018deep}. In particular, the introduction of transformers~\cite{2017Attention} has played a pivotal role in enabling high-performance trackers~\cite{TransT,Stark,wang2021transformer,mixformer,ostrack,SeqTrack,artrack}.
Despite these advancements, much of the research~\cite{zhao2022robust,SiameseRPN,DiMP,TransT,zhao2022vision,daivisual,liuspatial} has primarily focused on improving tracking accuracy, often overlooking the importance of speed. While many of these trackers can achieve real-time performance on high-end GPUs, their practical use is limited on resource-constrained devices. For instance, the high-performance tracker TransT~\cite{TransT} operates at only 5 frames per second (\emph{fps}) on an Intel Core i9-9900K CPU and 13 \emph{fps} on the NVIDIA Jetson AGX. This highlights the pressing need for a tracker that not only delivers high accuracy but also operates efficiently on devices with limited computational power.

\begin{figure}[t]
	\centering
	\subfloat[Comparing our HiT and DyHiT with other trackers on LaSOT.  ]{\label{fig:compare on last}\includegraphics[width=0.95\linewidth]{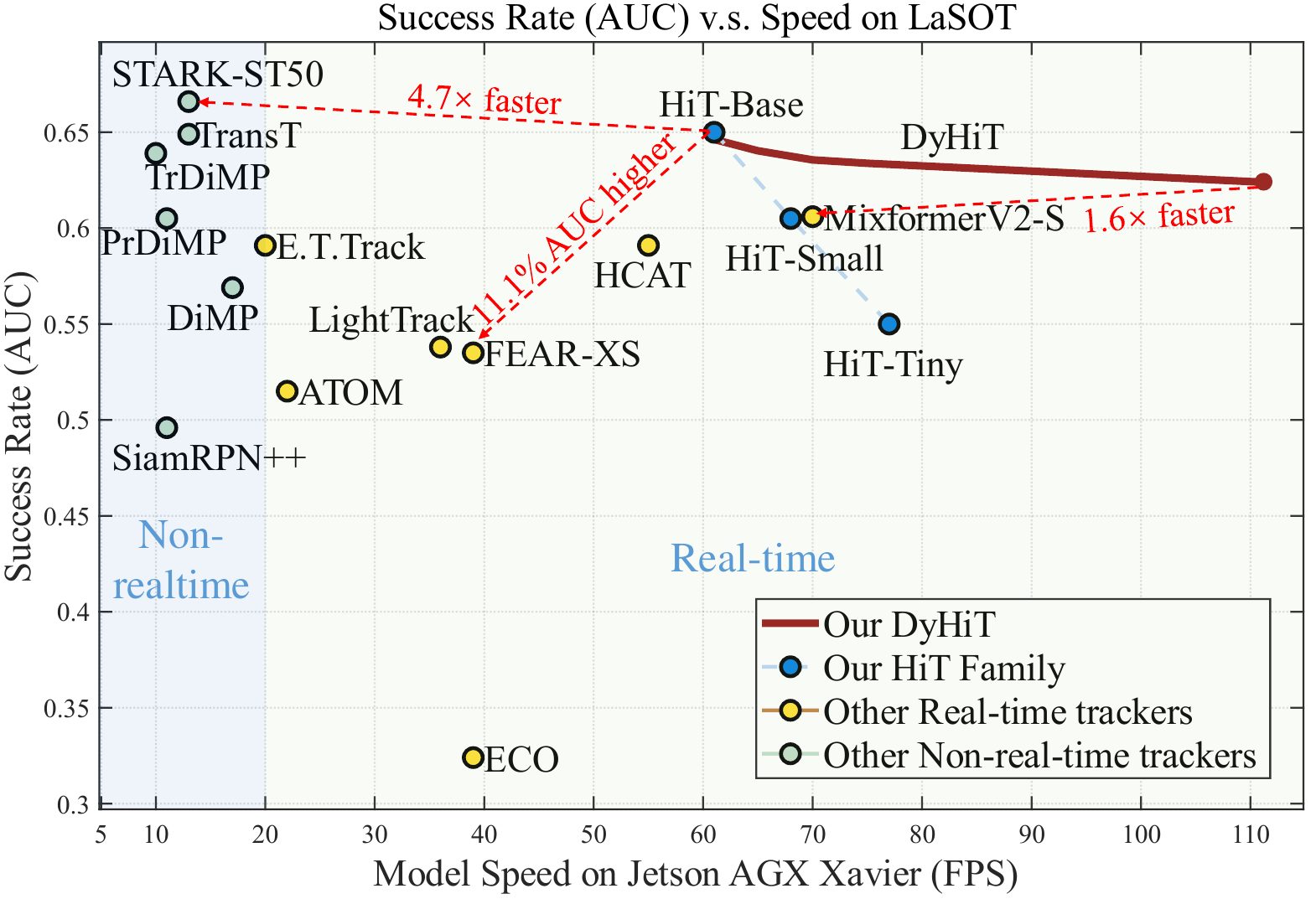}}\\
	\subfloat[Comparing the speed of DyTrackers and their base trackers.  ]{\label{fig:dytracker_speed2}\includegraphics[width=0.95\linewidth]{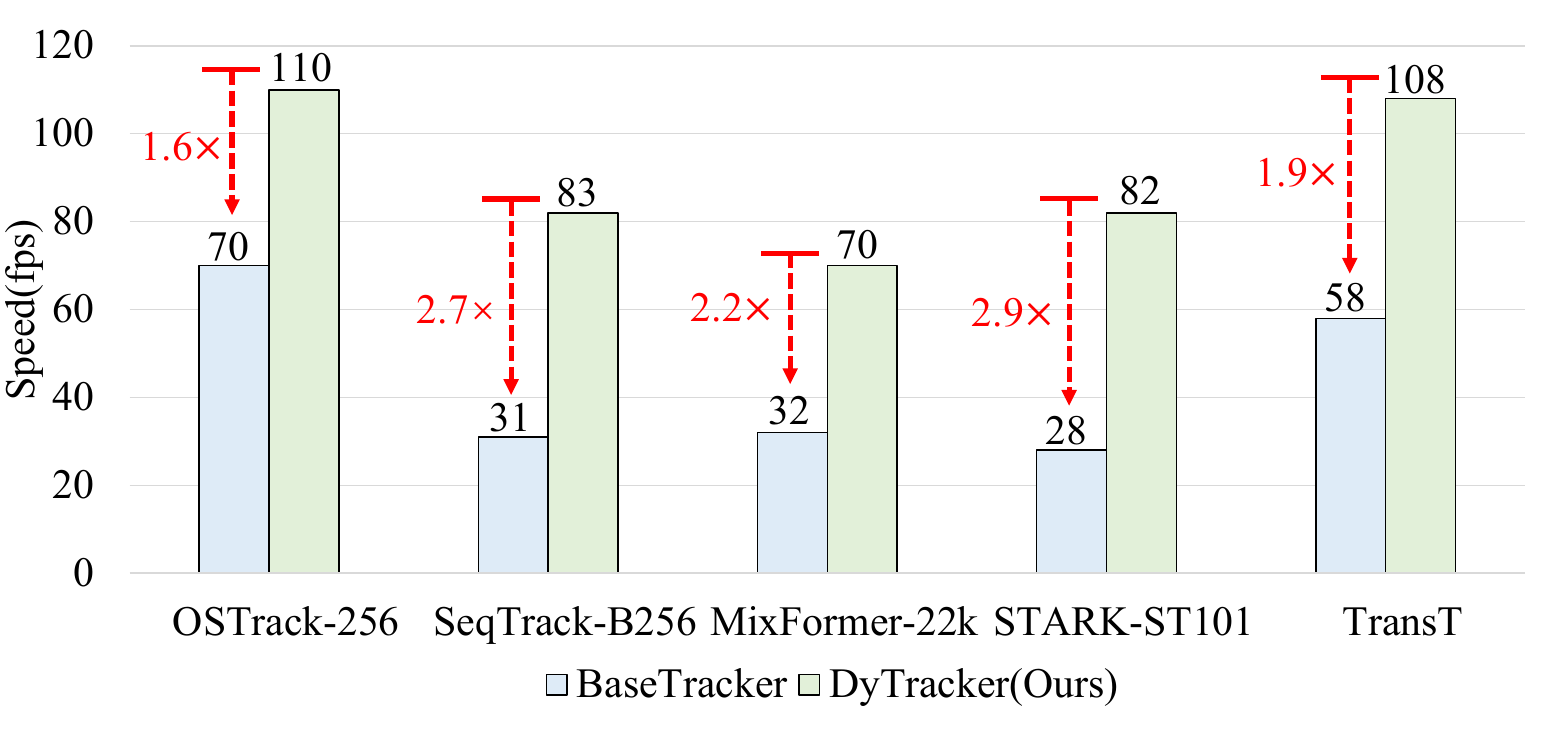}}\\	
	\caption{{Comparison of our methods and other trackers.  In (a), the reported speed refers to the inference speed of the tracker, excluding data pre-processing. Adhering to the VOT real-time setting~\cite{vot2020}, we set the real-time line at 20 \emph{fps} in (a).  For (b), we evaluate on the Nvidia GeForce RTX 2080 Ti GPU using a single thread on the GOT-10k test set. Subsequently, we submitted the test results to the GOT-10k evaluation server~\cite{GOT10K}, obtaining the speed.}}
\end{figure}

The one-stream structure has gained significant popularity in tracking applications~\cite{ostrack,simtrack,sbt,mixformer}. This architecture integrates feature extraction and feature fusion into a unified process, fully leveraging the potential of backbone networks~\cite{ViT} pre-trained for image classification. Following this trend, our study adopts the one-stream architecture, employing a lightweight transformer backbone network pre-trained for classification tasks.
However, a critical gap remains between the requirements of tracking and the design of networks optimized for image classification. Lightweight networks~\cite{graham2021levit,mehta2021mobilevit,wu2022tinyvit} in image classification typically adopt hierarchical architectures with high-stride downsampling to reduce computational costs. While this approach is effective in classification, the use of large-stride downsampling in tracking often leads to a loss of critical fine-grained information necessary for precise object localization. This discrepancy raises an question: How can we balance the need for detailed spatial information in tracking with the computational efficiency of high-stride downsampling in hierarchical backbone networks?

\begin{figure}[t]
\begin{center}
\includegraphics[width=1.0\linewidth]{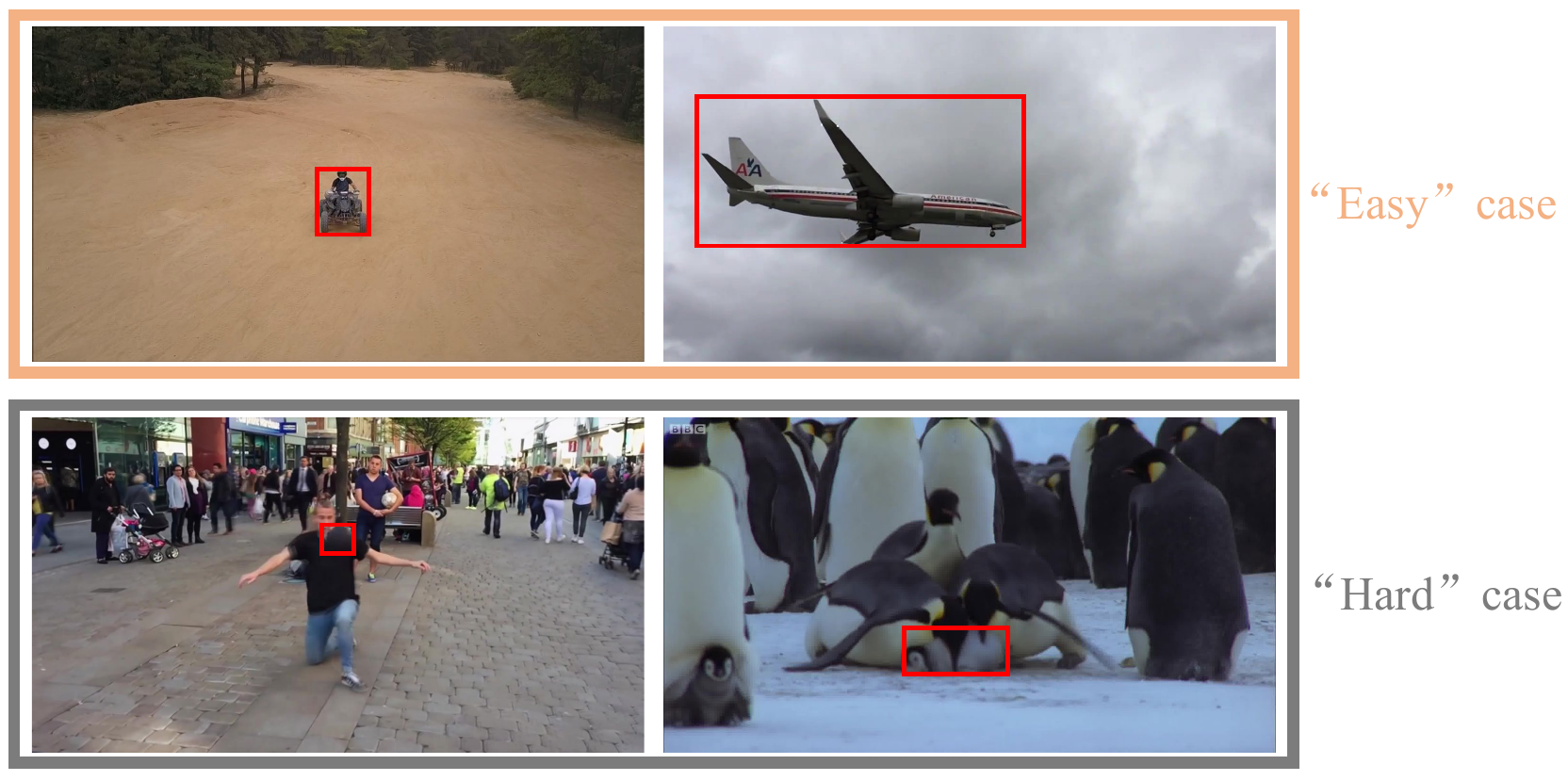}
\end{center}
   \caption{Examples of ``easy" and ``hard" cases.}
\label{fig:easy_hard}
\end{figure}

To address this challenge, we propose the Bridge Module, designed to integrate features from different stages of the hierarchical backbone. By merging deep semantic information with shallow, fine-grained details, the Bridge Module mitigates the information loss caused by large-stride downsampling. Incorporating this module into the lightweight hierarchical backbone LeViT~\cite{graham2021levit}, we develop HiT, a novel family of efficient tracking models.
In addition, we introduce a approach for relative position encoding, termed dual-image position encoding, to further enhance the representation of positional information. This method encodes the positional information of both the template and the search region simultaneously, promoting more effective interaction between the two and improving tracking performance.

Building upon the foundation of the HiT model family, we further propose an efficient dynamic tracking framework in this study. As illustrated in Fig.~\ref{fig:easy_hard}, tracking scenarios can range from simple to complex. Simple scenarios typically demand minimal computational resources and can be effectively handled by smaller models. In contrast, complex scenarios require larger models with greater computational capacity. However, conventional trackers rely on static models, applying the same processing framework to all scenarios. This results in inefficient resource utilization: over-provisioning for simple scenarios and inadequate performance for complex ones.
Previous studies~\cite{dytrack,east} have explored methods to partition tracking scenarios by incorporating dedicated judgment modules and sophisticated decision-making mechanisms. These methods activate different models based on scenario complexity, aiming to balance accuracy and efficiency. However, such approaches often introduce substantial latency due to their intricate judgment processes, making them unsuitable for efficient tracking. For example, the fastest version of preveious dynamic tracker~\cite{dytrack} achieves only 37 FPS on AGX, which limits its practical applications. Consequently, developing a lightweight and efficient mechanism to evaluate tracking scenarios and implement a divide-and-conquer strategy remains an urgent challenge.

To address this issue, we propose an efficient feature-driven dynamic routing architecture, extending our HiT model to implement DyHiT.
The core of this architecture lies in the incorporation of an early exit strategy into the framework. During forward propagation, the intermediate feature map of the search area is fed into a lightweight router to assess whether the current feature is sufficient to predict the tracking result accurately. If so, the forward propagation halts, and the intermediate feature is directly used for prediction. Otherwise, the propagation continues to extract more refined features.
This approach enables efficient adaptation to varying scenario complexities. In simple scenarios, only a shallow sub-network is activated, significantly reducing inference time. Conversely, in complex scenarios, deeper layers of the network are utilized to ensure precise predictions.
Unlike previous methods, our router is designed to be both simple and efficient, comprising only a few linear layers. This avoids the substantial latency typically introduced by the complex judgment modules and decision mechanisms used in earlier approaches. By leveraging feature-driven classification, our router performs straightforward assessments based on features extracted by the backbone, eliminating the need for intricate scene complexity evaluations.
As a result, DyHiT offers superior efficiency and practicality compared to prior dynamic trackers, achieving an optimal balance between speed and accuracy in resource-constrained environments.

Our comprehensive experiments demonstrate the effectiveness and efficiency of both HiT and DyHiT. As shown in Fig.\ref{fig:compare on last}, HiT-Base achieves an 11.1\% higher AUC score on the LaSOT benchmark compared to the high-speed tracker FEAR\cite{lin2017feature}, while operating at 1.6 times faster speed on the Nvidia Jetson AGX Xavier. Furthermore, when compared to the high-performance tracker STARK-ST50~\cite{Stark}, HiT-Base delivers comparable accuracy but with an impressive 4.7 times faster speed on the AGX, marking a substantial improvement over prior real-time trackers. 
The red line in Fig.\ref{fig:compare on last} illustrates the speed-accuracy trade-off curve of DyHiT, which achieves a broad spectrum of trade-offs using a single model. Notably, DyHiT outperforms all previous efficient trackers in both speed and precision. The fastest version of DyHiT operates at 111 FPS on the Nvidia Jetson AGX while maintaining an AUC of 62.4\% on LaSOT. This surpasses the recent MixformerV2-S\cite{mixformerv2} by 1.8\% in AUC, while achieving a speed that is 1.6 times faster on the AGX.

Moreover, building on the dynamic routing architecture of DyHiT, we introduce a training-free acceleration method for existing high-performance trackers. This approach significantly enhances tracking speed while maintaining accuracy.
We integrate the fastest route of DyHiT (DyHiT-Route1) and our efficient feature-driven router into various high-performance trackers to construct their corresponding fast variants, namely DyTracker.
During inference, DyHiT-Route1 first extracts image features, which are then evaluated by the router for reliability. If the features are deemed reliable, DyHiT-Route1 directly predicts the tracking results. Otherwise, the high-performance tracker is activated to ensure higher accuracy. This dynamic mechanism enables DyTracker to efficiently handle simple scenarios using DyHiT-Route1, conserving computational resources, while employing the high-performance tracker for complex scenarios to achieve precise predictions. As illustrated in Fig.\ref{fig:dytracker_speed2}, this approach delivers a substantial speed-up for existing high-performance trackers without compromising accuracy. For instance, state-of-the-art trackers OSTrack-256\cite{ostrack} and SeqTrack-B256~\cite{SeqTrack} achieve speed improvements of  $1.6\times $  and  $2.7\times $ , respectively, when augmented with our method.

Our main contributions can be summarized as:
\begin{itemize}[leftmargin=0.468cm]
\item{We introduce a new family of efficient tracking models, HiT. The proposed Bridge Module integrates high-level semantic information with shallow, fine-grained details, enabling the use of large-stride downsampling backbones in tracking. To improve positional accuracy, we introduce a dual-image position encoding approach that jointly encodes positional information from both the template and search region. HiT achieves superior performance and exceptional speed compared to previous efficient tracking methods.}

\item{We propose an efficient feature-driven dynamic routing architecture to extend HiT, resulting in DyHiT. DyHiT efficiently evaluates tracking scenarios and applies different routes based on their demands. A divide-and-conquer strategy is employed to maximize computational resource utilization. This approach achieves a wide range of speed-accuracy trade-offs.}
\item{Expanding upon DyHiT, we further develop a training-free acceleration method for high-performance trackers. Experiments on seven tracking models demonstrate that our method significantly improves the speed of high-performance trackers without compromising accuracy.}
\end{itemize}

This study builds on our conference paper~\cite{HiT} , which was published at the ICCV 2023 conference. We significantly extends it in various aspects. 
First, we extend HiT by incorporating the proposed efficient feature-driven dynamic routing architecture, implementing DyHiT. DyHiT achieves a wide range of speed-precision trade-offs, outperforming all previous efficient trackers. 
Second, we introduce a training-free acceleration method for existing high-performance trackers. This method significantly improves the speed of high-performance trackers without compromising their accuracy.
Third, we conduct a more comprehensive comparison of HiT and DyHiT with a variety of trackers on a broader range of datasets.

\section{Related Work}\label{sec:related work}
\subsection{Visual Tracking}\label{Visual Tracking}
Siamese-based methods~\cite{SiameseFC,SINT,SiameseRPN,SiamMask,unsupervised,SiamFC++,SiamCAR,SiamBAN,Deeper-wider-SiamRPN} have gained popularity in tracking. Typically, the Siamese-based framework employs two backbone networks with shared parameters to extract features from template and search region images. It utilizes a correlation-based network for feature interaction and head networks for final prediction. The introduction of transformers~\cite{2017Attention} in works like TransT~\cite{TransT}, TMT~\cite{wang2021transformer}, and their subsequent iterations~\cite{Dehazing,liu2021swin,ToMP,CSWinTT,AiATrack} further enhances tracking performance through advanced feature interaction. Recently, a one-stream framework has established new state-of-the-art performance in tracking, exemplified by methods like MixFormer~\cite{mixformer}, SBT~\cite{sbt}, SimTrack~\cite{simtrack}, OSTrack~\cite{ostrack}, and SeqTrack~\cite{SeqTrack}. This one-stream framework jointly performs feature extraction and feature fusion with the backbone network, proving simple yet effective by leveraging the capabilities of a pre-trained backbone. However, these methods are designed for powerful GPUs, and their speeds on edge devices are suboptimal, limiting their applicability. In this work, we focus on enhancing the efficiency of the one-stream framework, thereby broadening its applicability to a wider range of real-world scenarios.

\subsection{Efficient Tracking Network}\label{Efficient Tracking Network}
 Practical applications necessitate efficient trackers capable of achieving both high performance and fast speed on edge devices. Early methods such as ECO~\cite{danelljan2017eco} and ATOM~\cite{ATOM} achieve real-time speed on edge devices, but their performance lags behind current state-of-the-art trackers. Recently, some efficient trackers have emerged. LightTrack~\cite{yan2021lighttrack} employs NAS to search networks, resulting in low computational requirements and relatively high performance. FEAR~\cite{borsuk2022fear} achieves a family of efficient and accurate trackers by employing a dual-template representation and a pixel-wise fusion block. Despite these advances, a significant performance gap remains between these efficient trackers and high-performance trackers~\cite{TransT,ostrack}. In this work, we propose HiT and its extension, DyHiT. Both HiT and DyHiT achieve high speeds on edge devices while delivering competitive performance compared to high-performance trackers.

 \subsection{Vision Transformer}\label{Vision Transformer}
 ViT~\cite{ViT} introduces the transformer to image classification and demonstrates impressive performance. Subsequently, numerous vision transformer networks~\cite{touvron2021training,yuan2021tokens,wu2021cvt,wang2021pyramid,liu2021swin,eatformer} have been developed. While transformers are renowned for their superior modeling capabilities, their speed is a limitation. Hence, lightweight vision transformers~\cite{mehta2021mobilevit,graham2021levit,wu2022tinyvit} have emerged, significantly accelerating the speed of transformer-based networks. These lightweight transformers deviate from classical vision transformers by adopting a hierarchical architecture with high-stride downsampling, reducing computational overhead. In this work, we integrate a lightweight hierarchical vision transformer into the one-stream tracking framework, using LeViT~\cite{graham2021levit} as the default backbone. However, our approach differs fundamentally from LeViT in several key aspects: i) While LeViT makes predictions based on heavily downsampled features, our HiT framework incorporates a Bridge Module to fuse features from multiple stages, enabling predictions on fused high-resolution features. Additionally, we modify the transformer module to process both the search region and the template simultaneously. Our dynamic framework further enhances model efficiency. ii) LeViT is designed specifically for image classification, focusing more on high-level semantic information, while shallow details are less important. In tracking tasks, however, shallow details are crucial for accurate localization. Our model combines both high-level semantic information and shallow details, making it more suitable for tracking tasks. iii) LeViT employs position encoding for individual images. In comparison, we introduce dual-image position encoding to jointly represent positional information for both the template and the search region, enhancing the model's ability to capture fine details.

  \subsection{Dynamic network}\label{Dynamic network}
  In image classification, dynamic networks can be broadly categorized into two types: instance-wise~\cite{dydense,lidynamic,skipnet,wang2021not,radapt} and spatial-wise~\cite{SACT,spatial-wise,dynamicvit,ddpnas}. Spatial-wise dynamic networks, such as SACT~\cite{SACT} and the dynamic token sparsity framework~\cite{dynamicvit}, enhance efficiency by dynamically adjusting the execution layers of the network or pruning redundant tokens based on input characteristics, thereby accelerating processing speed.
  However, the performance of spatial-wise dynamic networks often depends on the co-design of software and hardware for optimized execution. In contrast, instance-wise dynamic networks are more adaptable to traditional CPUs and GPUs as they do not require sparse computation. A notable example is MSDNet~\cite{dydense}, which employs an early-exit strategy by training multiple classifiers with varying resource demands and integrating them into a unified neural network. During inference, these classifiers are adaptively utilized based on prediction confidence, enabling efficient acceleration.
  Dynamic networks have also been applied in visual tracking. For instance, EAST~\cite{east} formulates the adaptive tracking problem as a decision-making process using reinforcement learning and achieves dynamic tracking through the design of complex decision schemes. Similarly,  Zhu et al.~\cite{dytrack} introduce a dedicated and complex module for scene assessment, enabling the selection of appropriate inference paths for different inputs to achieve dynamic inference. However, the reliance on intricate judgment modules and decision schemes in previous works introduces significant time overhead for scene assessment, limiting their practical applications.
  In this work, we propose an efficient feature-driven dynamic routing architecture to extend our HiT, resulting in DyHiT. DyHiT enables rapid scene assessment and implements a divide-and-conquer strategy, ensuring both efficiency and practicality.

  \section{Method}\label{sec:Method}
This section presents a comprehensive overview of our model. We begin with a brief introduction to the HiT framework, followed by a detailed explanation of its architecture. This includes a lightweight hierarchical backbone with dual-image position encoding, a Bridge Module, and a tracking head. Next, we introduce DyHiT, our dynamic tracking extension, along with the acceleration strategy for high-performance trackers. Finally, we describe the training pipelines.

\begin{figure*}[t]
\begin{center}
\includegraphics[width=1\linewidth]{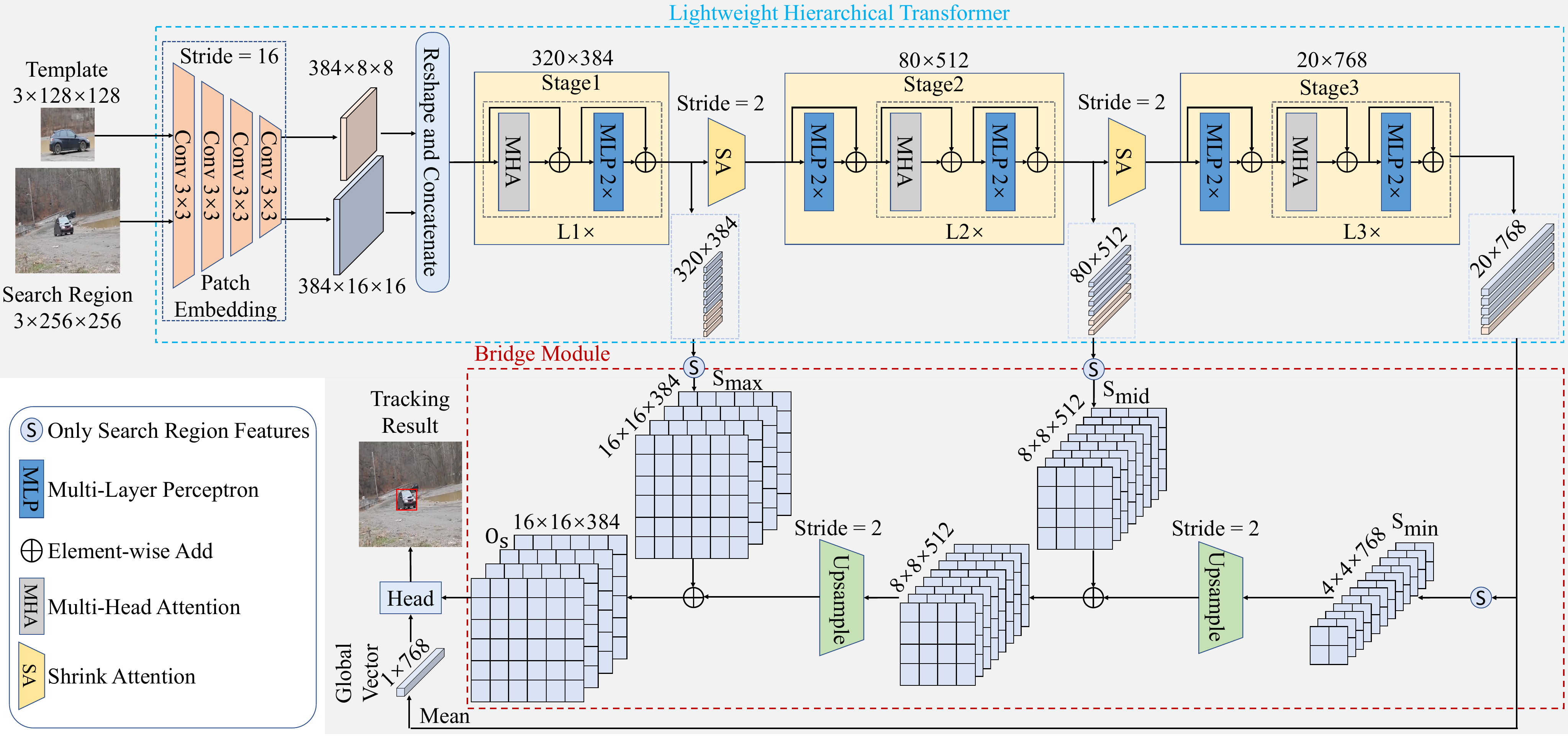}
\end{center}
   \caption{Architecture of the proposed HiT framework. The HiT framework contains three components: a lightweight hierarchical vision transformer, a Bridge Module, and a prediction head. }
\label{fig:framework}
\end{figure*}

\subsection{Overview}\label{sec-overview}
As depicted in Fig.~\ref{fig:framework}, HiT is a one-stream tracking framework comprising three key components: the lightweight hierarchical transformer, the proposed Bridge Module, and the head network. The input comprises an image pair, including the search region and template images, which is processed by the lightweight hierarchical transformer for simultaneous feature extraction and fusion. The core elements of the hierarchical vision transformer include Multi-Head Attention (MHA), Shrink Attention (SA), and dual-image position encoding. MHA extracts and fuses features from the search and template images, SA reduces the feature resolution for computational efficiency, and dual-image position encoding jointly encodes positional information for both images. Each stage of the transformer generates a sequence of features at different resolutions, culminating in a global vector obtained by averaging the final output features from the last stage. Subsequently, the feature sequence enters the Bridge Module, where features are fused to acquire enhanced features. Finally, the global vector and the enhanced features are fed into the prediction head to produce the tracking result.

\subsection{Lightweight Hierarchical ViT}
\textbf{Hierarchical Backbone.} We employ LeViT~\cite{graham2021levit} as the backbone of HiT, and adapt it into our tracking framework. Specifically, the backbone takes as input the template image $\mathbf{Z} \in {\mathbb{R}}^{3 \times {H_{z}} \times {W_{z}}}$ and the search region image $\mathbf{X} \in {\mathbb{R}}^{3 \times {H_{x}} \times {W_{x}}}$. Initially, we downsample the image pair by a factor of 16 through patch embedding, resulting in $\mathbf{Z_{p}} \in {\mathbb{R}}^{C \times {\frac{H_{z}}{16}} \times {\frac{W_{z}}{16}}}$ and $\mathbf{X_{p}} \in {\mathbb{R}}^{C \times {\frac{H_{x}}{16}} \times {\frac{W_{x}}{16}}}$. Subsequently, we flatten and concatenate $\mathbf{Z_{p}}$ and $\mathbf{X_{p}}$ in the spatial dimension before feeding them into the hierarchical transformer. 
The hierarchical transformer comprises three stages, with the $i$$-$$th$ stage having $Li$ blocks ($L1$$=$$L2$$=$$L3$$=$$4$, by default). Each block includes a Multi-Head Attention and an MLP in the residual form. Shrink Attention modules connect each stage and downsample features by a factor of 4 in the spatial dimension. For the output features of each stage, we extract partial features corresponding to the search image. In the final stage, we average its output features to obtain a global vector $\mathbf{G}$. After the transformer backbone, we obtain a global vector $\mathbf{G} \in {\mathbb{R}}^{1 \times C_{min}}$ and a feature sequence with three feature maps of different sizes: $\mathbf{S_{max}} \in {\mathbb{R}}^{{H_{max}} \times W_{max} \times C_{max}}$, $\mathbf{S_{mid}} \in {\mathbb{R}}^{{H_{mid}} \times W_{mid} \times C_{mid}}$, $\mathbf{S_{min}} \in {\mathbb{R}}^{{H_{min}} \times W_{min} \times C_{min}}$.

\begin{figure}[t]
	\centering
	\subfloat[Multi-Head Attention (MHA)]{\label{fig:MHA}\includegraphics[width=0.6\linewidth]{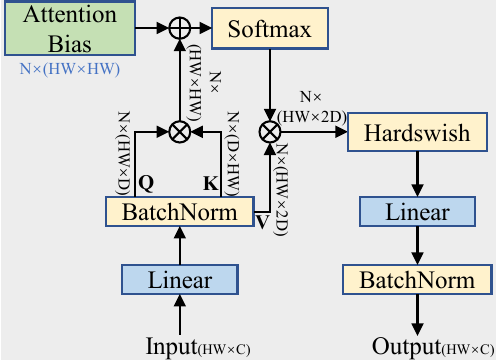}}\\
	\subfloat[Shrink Attention (SA)]{\label{fig:Shrink}\includegraphics[width=0.8\linewidth]{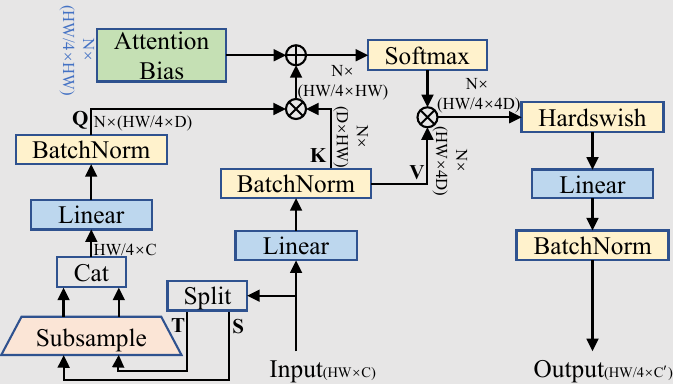}}\\	
	\caption{Detailed architectures of MHA and SA.}
\end{figure}

\noindent\textbf{Multi-Head Attention (MHA).} The structure of MHA is illustrated in Fig.~\ref{fig:MHA}. The number of channels of \textbf{Q} and \textbf{K} is half of \textbf{V} to reduce the amount of calculation. Following LeViT, we use attention bias as a relative position encoding rather than absolute position encoding. We generate the attention bias using our dual-image position encoding, the details of which will be introduced later. The mechanism of MHA can be summarized as:

\begin{equation}
\label{eq-m-h-a}
\begin{split}
     {\rm{Attn}}(\mathbf{Q},\mathbf{K},\mathbf{V},\mathbf{B}_i)
     = {\rm{softmax}}(\frac{\mathbf{Q}\mathbf{K}^\top}{\sqrt{d_k}}+\mathbf{B}_i)\mathbf{V}, \\
     {\mathbf{H}_i}={\rm{Hardswish}(\rm{Attn}}(\mathbf{X}\mathbf{W}_i^Q,\mathbf{X}\mathbf{W}_i^K,\mathbf{X}\mathbf{W}_i^V,\mathbf{B}_i)),\\
     {\rm{MultiHead}}(\mathbf{X}) = {\rm{Concat}}({\mathbf{H}_1},...,{\mathbf{H}_{N}}){\mathbf{W}^O}, 
\end{split}
\end{equation}
where $\mathbf{X}\in\mathbb{R}^{HW \times C}$ is the input, $\mathbf{B}_i\in\mathbb{R}^{HW \times HW}$ is the attention bias, and $\mathbf{W}_i^Q, \mathbf{W}_i^K \in \mathbb{R}^{C \times D}$, $\mathbf{W}_i^V \in \mathbb{R}^{C \times 2D}$, and $\mathbf{W}^O \in \mathbb{R}^{2ND \times C}$ are parameter matrices.

\noindent\textbf{Shrink Attention (SA).} The structure of SA is illustrated in Fig.~\ref{fig:Shrink}. SA plays a crucial role in connecting the stages of the hierarchical transformer and downsampling the features. The architecture of SA mirrors that of MHA with some notable modifications: 1) The 2D input features are split into template features (T) and search region features (S) based on their spatial positions. These features are then reshaped into 3D, subsampled by a factor of 2 in each spatial direction, and re-flattened before being concatenated along the spatial dimension. This process effectively reduces the size of \textbf{Q} by a factor of 4 overall, resulting in downsampled SA output.  2) To address potential information loss due to downsampling, the number of channels in \textbf{V} is doubled. Additionally, the output features are configured with an increased number of channels to enhance their representational capacity.
These modifications enable SA to efficiently downsample features while preserving critical information, ensuring effective integration within the hierarchical transformer framework.

\begin{figure}[t]
	\centering
 	\subfloat[Previous Position Encoding]{\label{fig:PE-previous}\includegraphics[width=0.9\linewidth]{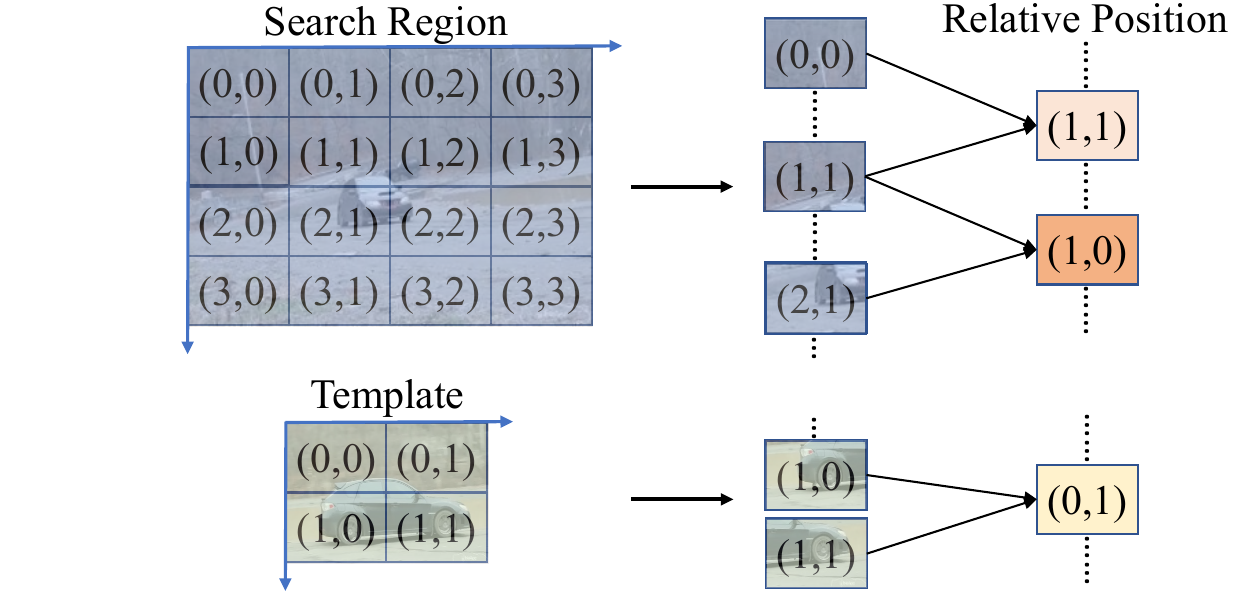}}\\	
	\subfloat[Our Dual-Image Position Encoding]{\label{fig:PE-our}\includegraphics[width=0.9\linewidth]{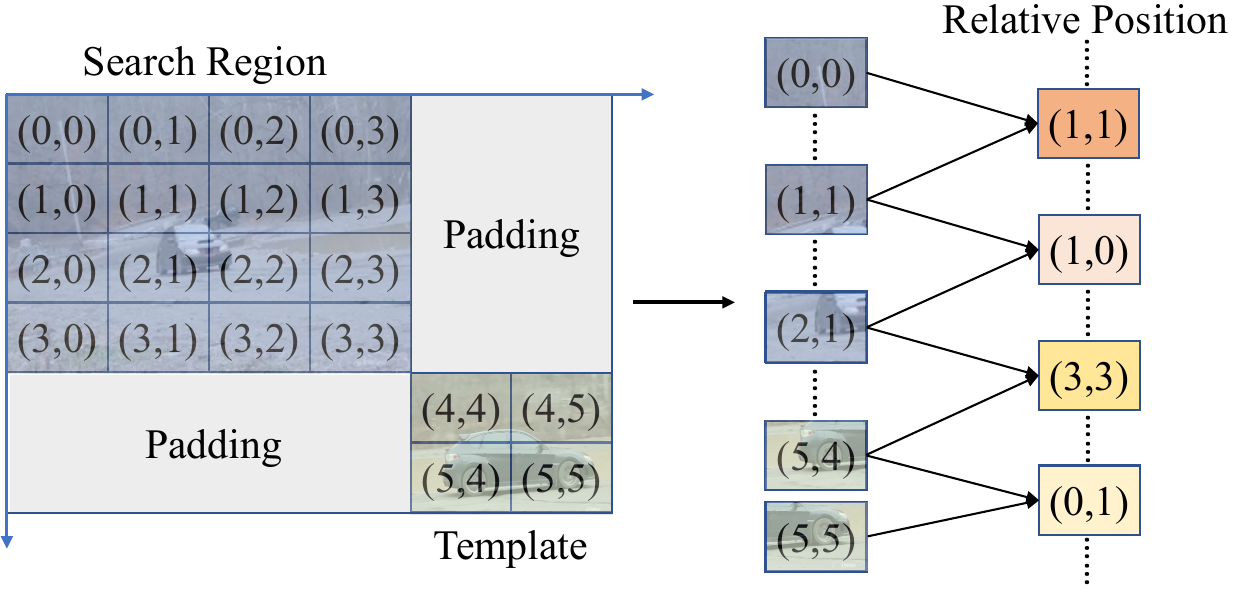}}\\
	\caption{Comparison of our dual-image position encoding and the previous position encoding.}
\end{figure}

\label{sec:PE}
\noindent\textbf{Dual-image Position Encoding.}  In line with LeViT, we utilize attention bias to incorporate relative position information into attention maps. To more effectively encode the joint position information of both the template and the search region, we propose the dual-image position encoding method. Specifically, attention bias is represented as a set of learnable parameters. The process involves computing the relative positions between every pair of pixels, utilizing these relative positions as indices to retrieve the corresponding learned parameters, and subsequently adding them to the attention map. This method effectively introduces crucial position information into the attention mechanism. It is calculated as 

\begin{equation}
\label{pos_encode}
\begin{array}{cc}
     {\rm{Bias}}^h = {\mathbf{B}^{h}(\lvert {x}-{x}^{'} \rvert, \lvert {y}-{y}^{'} \rvert)} 
\end{array}, 
\end{equation}
where $(x,y)$ and $({x}^{'},{y}^{'})$ $\in [H] \times [W]$ are the two pixels on the feature map. $\mathbf{B}^{h}$ is the learned parameters, and ${\rm{Bias}}^{h}$ is the indexed learned parameters. As illustrated in Fig.~\ref{fig:PE-previous}, the previous position encoding approach encodes the template and the search region separately. However, the positions of the two images partially overlap, leading to information confusion. Specifically, the position of the template aligns with the upper-left portion of the search region. To mitigate this issue, our dual-image position encoding adopts a diagonal arrangement for the template and the search region, encoding their position information jointly, as depicted in Fig.~\ref{fig:PE-our}. This diagonal arrangement ensures the encoding of unique horizontal and vertical coordinates for each pixel in the template and search region, thereby avoiding the confusion of detailed position information.

\subsection{Bridge Module and Head}
\textbf{Bridge Module.} The Bridge Module serves to fuse features from different stages of the hierarchical transformer, producing an enhanced feature that combines both detailed and semantic information. It acts as a bridge between the lightweight hierarchical transformer and the tracking framework. To ensure model efficiency, we aim for the Bridge Module to have a minimal yet effective architecture.
The simplicity of its design leads to compelling results. In Fig.~\ref{fig:framework}, the red box illustrates the outputs of the transformer:  three 2D features with distinct sizes. We reshape these 2D features into 3D feature maps denoted as $\mathbf{S_{min}}$, $\mathbf{S_{mid}}$, and $\mathbf{S_{max}}$. The Bridge Module follows a simple procedure: first, $\mathbf{S_{min}}$ is upsampled and added to $\mathbf{S_{mid}}$. Next, the resulting feature is upsampled and combined with $\mathbf{S_{max}}$, yielding the final enhanced feature. For all upsampling operations, a transpose convolutional layer with a stride of 2 is used. The mechanism of the Bridge Module can be summarized as 
\begin{equation}
\begin{split}
    {\mathbf{O_\text{s}}}
    = {\mathbf{S_\text{max}}}+{\rm{Upsample}}({\mathbf{S_\text{mid}}}+{\rm{Upsample}}({\mathbf{S_\text{min}}})),
\end{split}
\label{eq-att}
\end{equation}
where $\mathbf{O_{s}} \in {\mathbb{R}}^{{H_{max}} \times W_{max} \times C_{max}}$ is the output of the Bridge Module; $\mathbf{S_{max}} \in {\mathbb{R}}^{{H_{max}} \times W_{max} \times C_{max}}$, $\mathbf{S_{mid}} \in {\mathbb{R}}^{{H_{mid}} \times W_{mid} \times C_{mid}}$ and $\mathbf{S_{min}} \in {\mathbb{R}}^{{H_{min}} \times W_{min} \times C_{min}}$ are feature maps output by the lightweight hierarchical transformer. The Bridge Module effectively integrates deep semantic information with shallow detail, mitigating information loss caused by large-stride downsampling.{
It introduces only 327M FLOPs and 2.6M parameters, accounting for merely 7.5\% and 6.3\% of the total network, respectively.}
Despite its minimalist design, the module consistently produces compelling results while maintaining high efficiency.

\noindent\textbf{Head.} We use the corner head~\cite{Stark} for prediction. First, the attention map between ${\mathbf{G}}$ and ${\mathbf{O_{s}}}$ is computed. Next, ${\mathbf{O_{s}}}$ is re-weighted using the attention map, allowing local features to be enhanced or suppressed based on global information. Finally, the re-weighted ${\mathbf{O_{s}}}$ is passed through a fully-convolutional network to produce the target coordinates.

\begin{figure}[t]
\begin{center}
\includegraphics[width=1\linewidth]{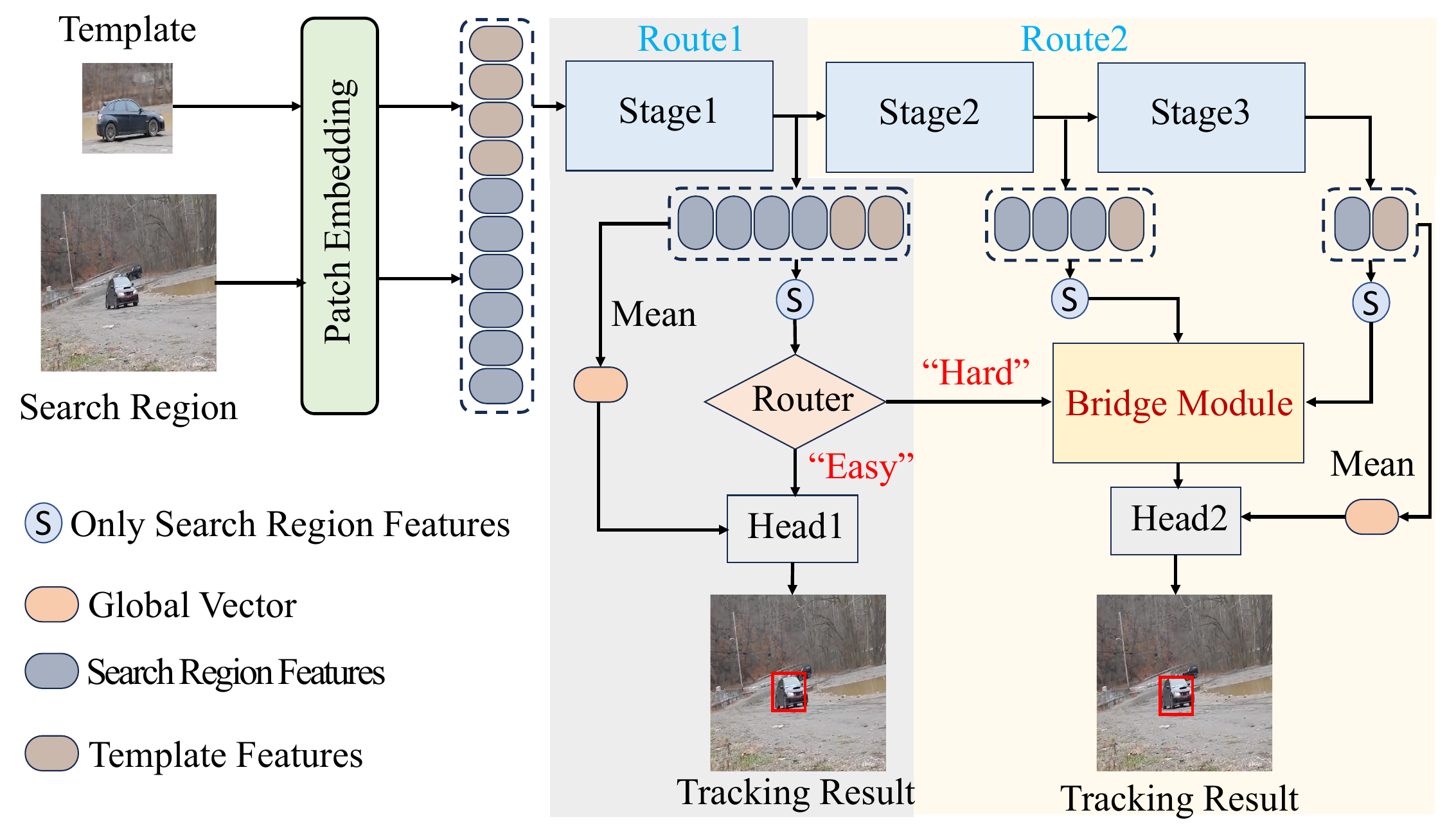}
\end{center}
   \caption{Framework of the proposed DyHiT. DyHiT consists of three components: a Router for assessing the complexity of scenes, Route1 for simple scenarios, and Route2 for complex scenarios.}
\label{fig:dyhit_framework}
\end{figure}
\subsection{Dynamic Routing Mechanism for HiT} \label{dyhit_scrib}
{To further improve the efficiency of HiT, we develop DyHiT by introducing an efficient feature-driven dynamic routing architecture to HiT-Base.}
DyHiT effectively classifies tracking scenarios and flexibly invokes different routes based on scene complexity, optimizing computational resource usage. This enables a wide range of speed-accuracy trade-offs with a single tracker.
As shown in Fig.~\ref{fig:dyhit_framework}, DyHiT and HiT share the same backbone network. The input images undergo patch embedding for downsampling, and the concatenated tokens after downsampling are input into the subsequent backbone network for processing. We divide the backbone network of HiT into two paths: one for handling simple scenes, referred to as Route1 in Fig.~\ref{fig:dyhit_framework}, and another for handling complex scenes, referred to as Route2. From the features $\mathbf{S_{1}}$ output by the first stage of the backbone, we extract the search region features ($\mathbf{S_{max}}$) and input them into the router. The router generates a score ($\mathbf{F}$), which is compared to a threshold ($\mathbf{T}$). If $\mathbf{F}$ exceeds $\mathbf{T}$, we classify the scene as simple, and the features from the first stage are sufficient for accurate prediction. In this case, we terminate the inference process and activate Route1 for prediction. The global vector $\mathbf{G}$ is obtained by computing the mean of $\mathbf{S_{1}}$, then $\mathbf{S_{max}}$ and $\mathbf{G}$ are fed into Head1 for prediction. If $\mathbf{F}$ is less than $\mathbf{T}$, we consider the scene to be complex and activate Route2. The backbone continues through stages 2 and 3, and the features from all three stages are fused via the Bridge Module. The fused features are then input into Head2 for prediction, resulting in the final tracking output.
\begin{figure}[t]
\begin{center}
\includegraphics[width=1\linewidth]{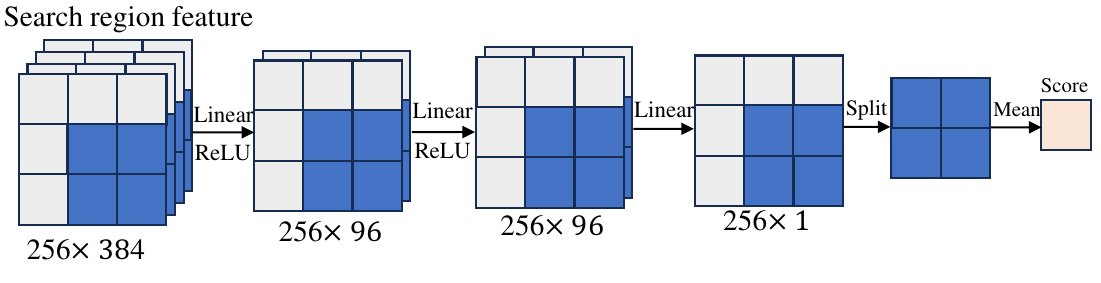}
\end{center}
   \caption{{Framework of the Router in DyHiT. The blue regions indicate the foreground, while the gray regions represent the background.}}
\label{fig:router}
\end{figure}
This divide-and-conquer strategy allows DyHiT to use shallow networks for fast predictions in simple scenarios, while invoking deeper networks for more precise predictions in complex scenarios. This optimizes computational resource utilization.
Both Head1 and Head2 utilize the corner head~\cite{Stark}, consistent with the head in HiT.
Notably, in our efficient feature-driven approach, the router only relies on the search region features extracted by the backbone, without additional feature extraction or integration. 
{As a result, the router design is extremely simple, as shown in Fig.~\ref{fig:router}, consisting of just three linear layers, which are sufficient for making accurate decisions based on the existing features.
The Router introduces only 11M FLOPs and 0.05M parameters, accounting for just 0.2\% and 0.1\% of the entire network, respectively, which is negligible.
}
This simplicity ensures the router’s high efficiency and avoids the substantial time costs typically associated with scene complexity assessments in previous methods.
In summary, DyHiT can be characterized as:

\begin{equation}
\label{eq-dyhit}
\begin{gathered}
     \mathbf{F} = \mathrm{Mean}(\mathrm{R}(\mathbf{S_\text{max}})), \\
     \mathbf{y} =
     \begin{cases}
         \mathrm{Head1}(\mathbf{S_\text{max}}, \mathbf{G}), & \text{if } \mathbf{F} > \mathbf{T} \\
         \mathrm{Head2}(\mathrm{B}(\mathbf{S_\text{max}}, \mathbf{S_\text{mid}}, \mathbf{S_\text{min}}), \mathbf{G}). & \text{if } \mathbf{F} < \mathbf{T}
     \end{cases}
\end{gathered}
\end{equation}
In the equation, $\mathrm{R}(\cdot)$ represents the router, consisting of three linear layers. $\mathrm{Mean}(\cdot)$ denotes the calculation of the mean, $\mathbf{F}$ signifies the difficulty score computed by the router, $\mathbf{T}$ is the threshold, $\mathbf{G}$ stands for the global vector, and $\mathbf{S_{max}}$, $\mathbf{S_{mid}}$, $\mathbf{S_{min}}$ represent the three feature maps output by the backbone network. $\mathrm{B}(\cdot)$ denotes the bridge module, and $\mathbf{y}$ represents the final prediction result. By adjusting $\mathbf{T}$, DyHiT can achieve a wide range of  speed-accuracy trade-offs.

\begin{figure}[t]
	\centering
 	\subfloat[Architecture of the high-performance tracker]{\label{fig:base_tracker}\includegraphics[width=0.68\linewidth]{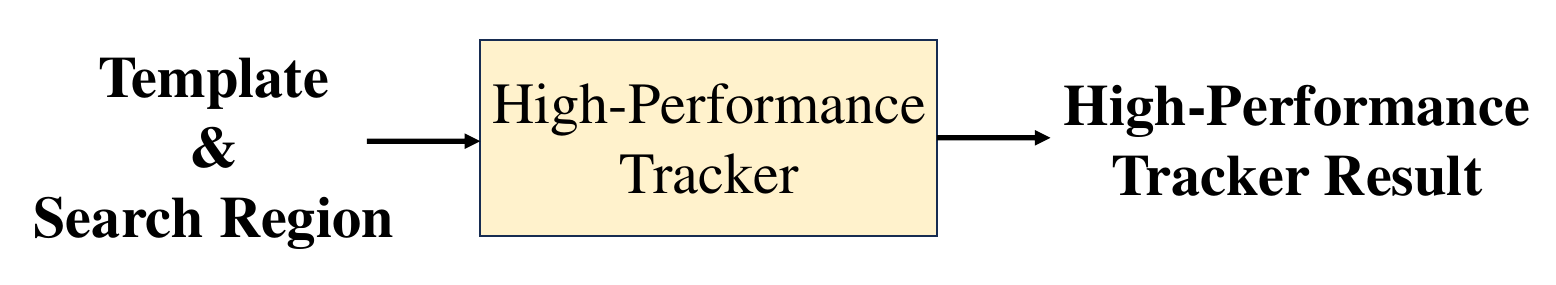}}\\	
	\subfloat[Architecture of our DyTracker]{\label{fig:dytracker}\includegraphics[width=0.9\linewidth]{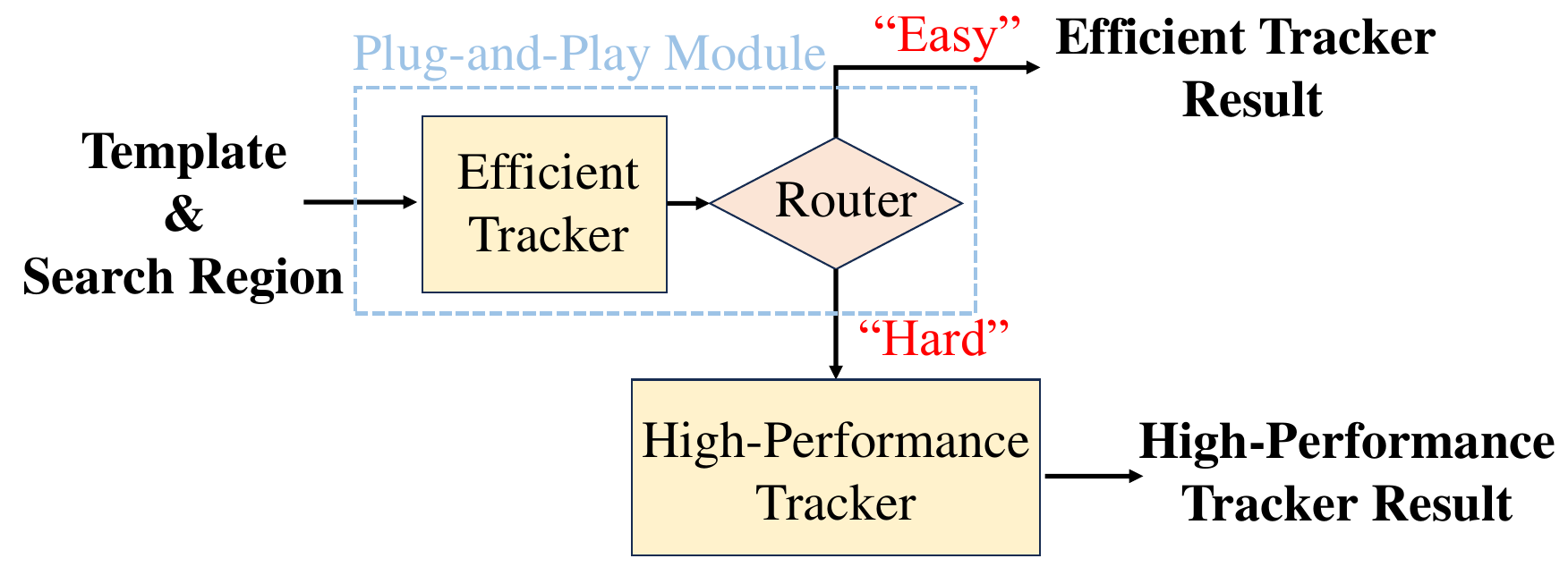}}\\
	\caption{Comparison of tracking frameworks. By combining the plug-and-play module with different trackers, we can obtain various DyTrackers.}
\end{figure}

\subsection{Training-Free Acceleration Method}
Based on the efficient dynamic routing architecture of DyHiT, we develop a training-free acceleration method to speed up the high-performance tracker. As shown in Fig.~\ref{fig:base_tracker}, existing high-performance trackers typically adopt a static structure, offering high accuracy but low speed. These trackers handle all tracking scenarios with a single model, which leads to inefficiencies since simple scenarios can often be accurately predicted using a lightweight model with minimal computational cost. This results in wasted computational resources when processing simple tracking scenarios. 
To address this issue, we propose a plug-and-play module, which can be flexibly integrated into existing high-performance base trackers without additional training process. By combining the plug-and-play module with different high-performance trackers, we can obtain various DyTrackers. As shown in Fig.~\ref{fig:dytracker}, the plug-and-play module consists of an efficient tracker and a router.
The efficient tracker utilizes the fastest route from DyHiT (Route1 in  DyHiT, denote as DyHiT-Route1), while the router design is identical to the one in DyHiT, ensuring efficient and consistent decision-making across different scenarios.
Firstly, we input the image pairs into DyHiT-Route1 for feature extraction. Subsequently, the extracted features are fed into the router for assessment. Similar to DyHiT, the router outputs a prediction score, and based on this score, we determine the difficulty level of the current scene for DyHiT-Route1. If it is deemed easy, we terminate the inference and directly use DyHiT-Route1 for prediction. Conversely, if it is considered challenging, we input the image pairs into the high-performance tracker for a re-prediction. 
This design enables a divide-and-conquer strategy: DyHiT-Route1 performs fast predictions in simple scenarios, while the high-performance tracker is activated for precise predictions in complex scenarios. This approach eliminates the computational resource waste typically associated with handling simple scenarios using a large model, while maintaining high accuracy in complex scenarios. 
{As a result, DyTracker enables the acceleration of high-performance trackers without sacrificing their accuracy.}

\subsection{Training Objective }\label{sec:training_and_inference}
\textbf{HiT.} For HiT, we combine the $\ell_1$ loss and the generalized GIoU loss~\cite{GIoU} as the training objective. The loss function is formulated as:
\begin{equation}
	\label{equ-loss-loc}
	\begin{aligned}
		\mathcal{L}=\lambda_\text{G}\mathcal{L}_\text{GIoU}(b_i,{\hat{b}}_i)+\lambda_\text{l}\mathcal{L}_\text{l}(b_i,\hat{b}_i),
	\end{aligned}
\end{equation}
where $b_i$ represents the ground-truth, and ${\hat{b}}_i$ represents the predicted box. $\lambda_{G}$ and $\lambda_{l}$ are weights, in experiments, we set $\lambda_{G}=2$ and $\lambda_{l}=5$. 

\noindent\textbf{DyHiT.} The training process of DyHiT consists of two stages. The objective of the first stage aligns with that of HiT, as illustrated in Equation~\ref{equ-loss-loc}. In the second stage, our aim is to enhance the router's ability to classify scenes. 
The feature map $\mathbf{F} \in {\mathbb{R}}^{{H_{x}} \times W_{x}\times C_{x}}$ is input into the router, producing the score map $\mathbf{S} \in {\mathbb{R}}^{{H_{x}} \times W_{x}}$ as output.
{We divide the score map $\mathbf{S}$ into positive and negative samples. Specifically, based on the ground truth bounding box, the score points located inside the ground truth bounding box in the score map $\mathbf{S}$ are classified as positive samples, while the other score points are classified as negative samples.}
All samples contribute to the overall loss. The loss in the second stage comprises three components: $\ell_1$ loss, GIoU loss, and MSE loss. The overall loss can be summarized as:
\begin{equation}
	\label{equ-loss-dyhit}
	\begin{gathered}
		\mathcal{L} = \lambda_\text{G}\mathcal{L}_\text{GIoU}(b_i,{\hat{b}}_i)+\lambda_\text{l}\mathcal{L}_\text{l}(b_i,\hat{b}_i)+\lambda_\text{R}\mathcal{L}_\text{MSE}(y_i,{\hat{y}}_i),\\
        \mathcal{L}_\text{MSE}(y_i,{\hat{y}}_i) = \frac{1}{n} \sum_{i=1}^{n} (y_i - \hat{y}_i)^2,
	\end{gathered}
\end{equation}
where $b_i$ represents the ground-truth, and ${\hat{b}}_i$ represents the predicted box. $y_i$ denotes the ground-truth label of the  $i$-th sample, when a sample belongs to the positive class, $y_i$ is equal to the intersection over union (IoU) value calculated between the corresponding $b_i$ and ${\hat{b}}_i$. When a sample is a negative example, $y_i$ = 0. ${\hat{y}}_i$ represents the score predicted by the router for the $i$-th sample. $\lambda_{G}$, $\lambda_{l}$ and $\lambda_{R}$ are weights, in experiments, we set $\lambda_{G}=\lambda_{l}=1$ and $\lambda_{R}=5$. 

\section{Experiments}
\subsection{ Implementation Details}\label{implementation}
\textbf{Model.} We develop three HiT variants with different lightweight transformers, as shown in Table~\ref{tab-modelf}. We adopt LeViT-384~\cite{graham2021levit}, LeViT-128, and LeViT-128S for HiT-Base, HiT-Small, and HiT-Tiny, respectively. Table~\ref{tab-modelf} also reports model parameters, MACs, and inference speed on multiple devices. All models are implemented with Python 3.8 and PyTorch 1.11.0. 

\begin{table}[t]\footnotesize
  \centering
    \caption{{Details of our HiT model variants.}}
  \label{tab-modelf}
  \setlength{\tabcolsep}{2mm}{  
  \begin{tabular}{c | l | c c c c c c}
    \toprule
     \multicolumn{2}{c|}{Model}&HiT-Base &&HiT-Small &&HiT-Tiny\\
    \midrule[0.5pt]
     \multirow{2}*{{PyTorch}}&GPU &175&&192&&204\\
    \multirow{3}*{{Speed (\emph{fps})}}&CPU &33&&72&&76\\
    &AGX &61&&68&&77\\
    &NX &32 &&34 &&39 \\
    \midrule[0.5pt]
    \multirow{2}*{{ONNX}}&GPU &274&&400&&455\\
    \multirow{2}*{{Speed (\emph{fps})}}&CPU &42&&98&&125\\
    &AGX &75&&119&&145\\
    \midrule[0.5pt]
    \multicolumn{2}{c|}{Macs (G)} &4.34&&1.13&&0.99\\
    \midrule[0.5pt]
    \multicolumn{2}{c|}{Params (M)} &42.14&&11.03&&9.59\\
  \bottomrule
\end{tabular}
}
\end{table}
\noindent\textbf{Training for HiT.} The training datasets of HiT encompass the train-splits of popular datasets including TrackingNet~\cite{trackingnet}, GOT-10k~\cite{GOT10K}, LaSOT~\cite{LaSOT}, and COCO2017~\cite{COCO}. The network processes image pairs, each comprising a template image and a search image. In the case of video datasets, we randomly select image pairs from video sequences. For the COCO image dataset, a random image is sampled, and then data augmentations, such as scaling, translation, and jittering, are applied to create an image pair. To define the search region and template, we expand the target box by factors of $4$ and $2$, respectively. Subsequently, the search and template images are resized to $256 \times 256$ and $128 \times 128$, respectively. The transformer backbone is initialized with the pretrained LeViT~\cite{graham2021levit} from ImageNet~\cite{ImageNet}. The remaining model parameters are initialized randomly. We employ the AdamW optimizer~\cite{AdamW} with a weight decay of 1e-4 and an initial learning rate of 5e-4 for training. Our model is trained for 1500 epochs on 4 NVidia RTX 3090 GPUs, with a batch size of 128 and each epoch comprising 60,000 sampling pairs. Notably, the learning rate undergoes a $10\times$ reduction at epoch 1200.

\noindent\textbf{Training for DyHiT.} The training process for DyHiT consists of two distinct stages. Initially, we focus on training Route1 of DyHiT, and subsequently, we train the router responsible for evaluating scene difficulty. 
The dataset and preprocessing strategy used for training DyHiT are the same as those employed for HiT. During Route1 training, we initialize the backbone and Head2 of DyHiT with parameters from the pre-trained HiT model, while Head1 is initialized randomly. To prevent interference with Route2 predictions, we freeze the backbone network and parameters of Head2, and only update Head1. This ensures that Head1 makes accurate predictions based on the features obtained in the first stage. For this stage, we use the AdamW optimizer with a weight decay of 1e-4 and an initial learning rate of 1e-4. Training runs for 90 epochs on a single NVIDIA A100 GPU with a batch size of 128. Each epoch consists of 60,000 sample pairs, and the learning rate is reduced by a factor of 10 at epoch 70. At the end of this stage, accurate predictions from both Route1 and Route2 are obtained.
In the second stage, we freeze other parameters and train the router. This stage lasts 60 epochs, using a single NVIDIA A100 GPU, with a batch size of 128 and 60,000 sample pairs per epoch. As in the first stage, the learning rate is reduced by a factor of 10 at epoch 48, while the other parameters remain the same as in the initial stage.
For DyTracker, no additional training is required. We simply combine Route1 and the router from DyHiT with existing high-performance trackers to construct DyTrackers for inference.

\begin{table*}[t]\small
  \centering
  \caption{State-of-the-art comparison on TrackingNet~\cite{trackingnet}, LaSOT~\cite{LaSOT}, and GOT-10k~\cite{GOT10K} benchmarks. We use \textcolor{gray}{gray} color to denote our trackers. The best three real-time results are shown in \textbf{\textcolor{cRed}{red}}, \textcolor{blue}{blue} and \textcolor{cGreen}{green} fonts, and the best non-real-time results are shown in \uline{\emph{underline}} font. We use $^\ast$ to indicate that  the results of the corresponding models on GOT-10k  are only trained with GOT-10k training set, while others are trained with additional datasets. {The reported speed refers to the inference speed of the tracker, excluding data pre-processing.}}
  \label{tab-sota}
  \setlength{\tabcolsep}{1.25mm}{  
  \begin{tabular}{c|l|c ccc c ccc c ccc c ccc}
    \toprule
    & \multirow{2}*{Method} &  \multicolumn{3}{c}{TrackingNet}&& \multicolumn{3}{c}{LaSOT} &&  \multicolumn{3}{c}{GOT-10k} && \multicolumn{3}{c}{PyTorch Speed (\emph{fps})} \\
    \cmidrule{3-5}
    \cmidrule{7-9}
    \cmidrule{11-13}
    \cmidrule{15-17}
   && AUC&P$_{Norm}$&P && AUC&P$_{Norm}$&P  && AO&SR$_{0.5}$&SR$_{0.75}$ && GPU&CPU&AGX\\
    \midrule[0.5pt]
    \multirow{11}*{\rotatebox{90}{Real-time}}&\cellcolor{mygray1}HiT-Base&\cellcolor{mygray1}\textbf{\textcolor{cRed}{80.0}}&\cellcolor{mygray1}\textbf{\textcolor{cRed}{84.4}}&\cellcolor{mygray1}\textbf{\textcolor{cRed}{77.3}} &\cellcolor{mygray1} &\cellcolor{mygray1}\textbf{\textcolor{cRed}{64.6}}&\cellcolor{mygray1}\textbf{\textcolor{cRed}{73.3}}&\cellcolor{mygray1}\textbf{\textcolor{cRed}{68.1}} &\cellcolor{mygray1}&\cellcolor{mygray1}\textcolor{blue}{64.0}&\cellcolor{mygray1}\textcolor{cGreen}{72.1}&\cellcolor{mygray1}\textbf{\textcolor{cRed}{58.1}} &\cellcolor{mygray1}&\cellcolor{mygray1}175&\cellcolor{mygray1}33&\cellcolor{mygray1}61\\
    &\cellcolor{mygray1}HiT-Small &\cellcolor{mygray1}\textcolor{cGreen}{77.7} &\cellcolor{mygray1}{81.9} &\cellcolor{mygray1}\textcolor{cGreen}{73.1} &\cellcolor{mygray1} &\cellcolor{mygray1}{60.5} &\cellcolor{mygray1}{68.3} &\cellcolor{mygray1}\textcolor{cGreen}{61.5} &\cellcolor{mygray1} &\cellcolor{mygray1}{62.6} &\cellcolor{mygray1}71.2 &\cellcolor{mygray1}{54.4} &\cellcolor{mygray1} &\cellcolor{mygray1}192 &\cellcolor{mygray1}72 &\cellcolor{mygray1}68  \\
    &\cellcolor{mygray1}HiT-Tiny &\cellcolor{mygray1}74.6 &\cellcolor{mygray1}78.1 &\cellcolor{mygray1}68.8 &\cellcolor{mygray1} &\cellcolor{mygray1}54.8 &\cellcolor{mygray1}60.5 &\cellcolor{mygray1}52.9 &\cellcolor{mygray1} &\cellcolor{mygray1}52.6 &\cellcolor{mygray1}59.3 &\cellcolor{mygray1}42.7 &\cellcolor{mygray1} &\cellcolor{mygray1}204 &\cellcolor{mygray1}76 &\cellcolor{mygray1}77  \\
    &\cellcolor{mygray1}DyHiT &\cellcolor{mygray1}\textcolor{blue}{77.9} &\cellcolor{mygray1}\textcolor{cGreen}{82.2} &\cellcolor{mygray1}\textcolor{blue}{73.8} &\cellcolor{mygray1} &\cellcolor{mygray1}\textcolor{blue}{62.4} &\cellcolor{mygray1}\textcolor{blue}{70.1} &\cellcolor{mygray1}\textcolor{blue}{64.0} &\cellcolor{mygray1} &\cellcolor{mygray1}\textcolor{cGreen}{62.9} &\cellcolor{mygray1}71.8 &\cellcolor{mygray1}\textcolor{cGreen}{55.2} &\cellcolor{mygray1} &\cellcolor{mygray1}299 &\cellcolor{mygray1}63 &\cellcolor{mygray1}111  \\
    &MixFormerV2-S~\cite{mixformerv2}& 75.8&81.1&70.4 && \textcolor{cGreen}{60.6}&\textcolor{cGreen}{69.9}&60.4 &&  62.1&-&- && 299&47&70\\
    &FEAR~\cite{borsuk2022fear}& -&-&- && 53.5&-&54.5 &&  61.9&\textcolor{blue}{72.2}&- && 105&60&38\\
    &HCAT~\cite{chen2022efficient}& {76.6}&\textcolor{blue}{82.6}&{72.9} &&{59.3}&{68.7}&{61.0} &&  \textbf{\textcolor{cRed}{65.1}}&\textbf{\textcolor{cRed}{76.5}}&\textcolor{blue}{56.7} &&195&45&55 \\
    &E.T.Track~\cite{ETTrack}& 75.0&80.3&70.6 && 59.1&-&- &&  -&-&- &&40&47&20\\
    &LightTrack~\cite{yan2021lighttrack}& 72.5&77.8&69.5 && 53.8&-&53.7 &&  61.1&71.0&- && 128&41&36\\
    &ATOM~\cite{ATOM}& 70.3&77.1&64.8 && 51.5&57.6&50.5 &&  55.6&63.4&40.2 &&83&18&22 \\
    &ECO~\cite{danelljan2017eco}$^\ast$ & 55.4&61.8&49.2 && 32.4&33.8&30.1 &&  31.6&30.9&11.1 && 240&15&39 \\
    \midrule[0.5pt]
    
    \multirow{12}*{\rotatebox{90}{Non-real-time}}
    &MCITrack-B224~\cite{mcitrack}$^\ast$& \emph{\uline{86.3}}&\emph{\uline{90.9}}&\emph{\uline{86.1}} && \emph{\uline{75.3}} &\emph{\uline{85.6}} &\emph{\uline{83.3}} &&  \emph{\uline{77.9}}&\emph{\uline{88.2}}&76.8 &&34&-&6\\
    &SUTrack-B224~\cite{sutrack}$^\ast$& 85.7&90.3&85.1 && 73.2 &83.4 &80.5 &&  \emph{\uline{77.9}}&87.5&\emph{\uline{78.5}} &&55&6&13\\
    &MixFormerV2-B~\cite{mixformerv2}& 83.4&88.1&81.6 && 70.6 &80.8 &76.2 &&  73.9&-&-&&116&11&16\\
    &SeqTrack-B256~\cite{SeqTrack}$^\ast$& 83.3&88.3&82.2 && 69.9&79.7&76.3 &&  74.7&84.7&71.8 && 31&-&6 \\
    &ARTrack-256~\cite{artrack}$^\ast$& 84.2&88.7&83.5 && 70.4&79.5&76.6 &&  73.5&82.2&70.9 && 39&-&6 \\
    &OSTrack-256~\cite{ostrack}$^\ast$& 83.1&87.8&82.0 && 69.1&78.7&75.2 &&  71.0&80.4&68.2 && 105&11&19 \\
    
    &MixFormer-22k~\cite{mixformer}$^\ast$& 83.1&88.1&81.6 && 69.2&78.7&74.7 &&  70.7&80.0&67.8 &&40&-&13\\
    &Sim-B/16~\cite{simtrack}$^\ast$&82.3&86.5&-&&69.3&78.5&-&&68.6&78.9&62.4&&87&10&16\\
    &STARK-ST50~\cite{Stark}$^\ast$& 81.3&86.1&-  && 66.6&-&- && 68.0&77.7&62.3 && 50&7&13 \\
    &TransT~\cite{TransT}$^\ast$& 81.4&86.7&80.3 && 64.9&73.8&69.0 && 67.1&76.8&60.9 && 63&5&13  \\
    &TrDiMP~\cite{wang2021transformer}$^\ast$& 78.4&83.3&73.1 && 63.9&-& 61.4 &&  68.8&80.5&59.7 && 41&5&10 \\
    &TrSiam~\cite{wang2021transformer}$^\ast$& 78.1&82.9&72.7 && 62.4&-&60.6 &&  67.3&78.7&58.6 && 40&5&10 \\
    &PrDiMP~\cite{PrDiMP}$^\ast$& 75.8&81.6&70.4 && 59.8&68.8&60.8 &&  63.4&73.8&54.3 &&47&6&11 \\
    &DiMP~\cite{DiMP}$^\ast$& 74.0&80.1&68.7 && 56.9&65.0&56.7 && 61.1&71.7&49.2 && 77 &10&17  \\
    &SiamRPN++~\cite{SiamRPNplusplus}$^\ast$& 73.3&80.0&69.4 && 49.6&56.9&49.1 && 51.7&61.6&32.5 && 56 &4&11 \\
\bottomrule
\end{tabular}
}
\end{table*}

\noindent\textbf{Inference.} During the inference, we begin by initializing the template in the first frame of a video sequence. For each subsequent frame, the search region is cropped based on the bounding box of the target from the previous frame. HiT operates as an end-to-end framework, where both the template and search images are input into the tracker, and the output of the model provides the final result. No additional post-processing techniques, such as window penalty or scale penalty~\cite{SiameseRPN}, are employed.
{For DyHiT, before determining the prediction path, the model first employs a router to assess the difficulty level of each frame.}
After the first stage, features related to the search region are input into the router, which outputs 256 scores ranging from 0 to 1.
{We set a foreground-background segmentation threshold, with a default value of 0.6.}
Scores above the threshold are selected, and the average score is calculated to derive the output of the router. This score classifies the scene as either easy or difficult. If classified as easy, inference by the backbone network is terminated, and the prediction from Route1 is used; otherwise, the prediction from Route2 is used. 
DyTracker follows a similar approach. Initially, features from the efficient tracker are input into the router to obtain a difficulty score, which determines the complexity of the scene. For easy scenes, the efficient tracker handles predictions; for more complex scenes, the high-performance tracker is activated.

\subsection{State-of-the-Art Comparisons}
Based on the speed on the Nvidia Jetson AGX Xavier edge device, we categorize trackers into real-time and non-real-time trackers. Consistent with the VOT real-time setting~\cite{vot2020}, we define the real-time threshold at 20 \emph{fps}. The evaluation compares HiT and DyHiT with state-of-the-art trackers across both real-time and non-real-time categories, using seven tracking benchmarks. The speed assessments are conducted on three platforms: Nvidia GeForce RTX 2080 Ti GPU, Intel Core i9-9900K @ 3.60GHz CPU, and the Nvidia Jetson AGX Xavier edge device. Results are presented in Tables~\ref{tab-sota} and~\ref{tab-sota-small}.
{The results of DyHiT reported in Table~\ref{tab-sota} and Table~\ref{tab-sota-small} are based solely on Route1, without employing dynamic routing. For the performance with dynamic routing enabled, please refer to the results presented in Fig.~\ref{fig:threshold_influence}.}

\noindent\textbf{Speed.} Table~\ref{tab-sota} presents the speeds of various trackers. On the GPU, HiT-Base, HiT-Small, HiT-Tiny, and DyHiT operate at 175 \emph{fps}, 192 \emph{fps}, 204 \emph{fps}, and 299 \emph{fps}, respectively, showcasing speeds 1.66$\times$, 1.82$\times$, 1.94$\times$, and 2.85$\times$ faster than FEAR~\cite{borsuk2022fear}. On the AGX edge device, HiT-Base, HiT-Small, HiT-Tiny, and DyHiT achieve speeds of 61 \emph{fps}, 68 \emph{fps}, 77 \emph{fps}, and 111 \emph{fps}, respectively, surpassing FEAR by 1.61$\times$, 1.79$\times$, 2.03$\times$, and 2.92$\times$. On the CPU, HiT-Base, HiT-Small, HiT-Tiny, and DyHiT reach speeds of 33 \emph{fps}, 72 \emph{fps}, 76 \emph{fps}, and 63 \emph{fps}. While only HiT-Base is slightly slower than FEAR, it still maintains a real-time speed. {In addition, we also evaluate the speed of HiT on the NVIDIA Jetson Xavier NX. As shown in Table~\ref{tab-modelf}, HiT-Base, HiT-Small, and HiT-Tiny achieve speeds of 32fps, 34fps, and 39fps, respectively.
}
Overall, HiT and DyHiT exhibit impressive speeds across multiple devices, suggesting its suitability for various tracking applications.

\begin{figure}[t]
\begin{center}
\includegraphics[width=1.0\linewidth]{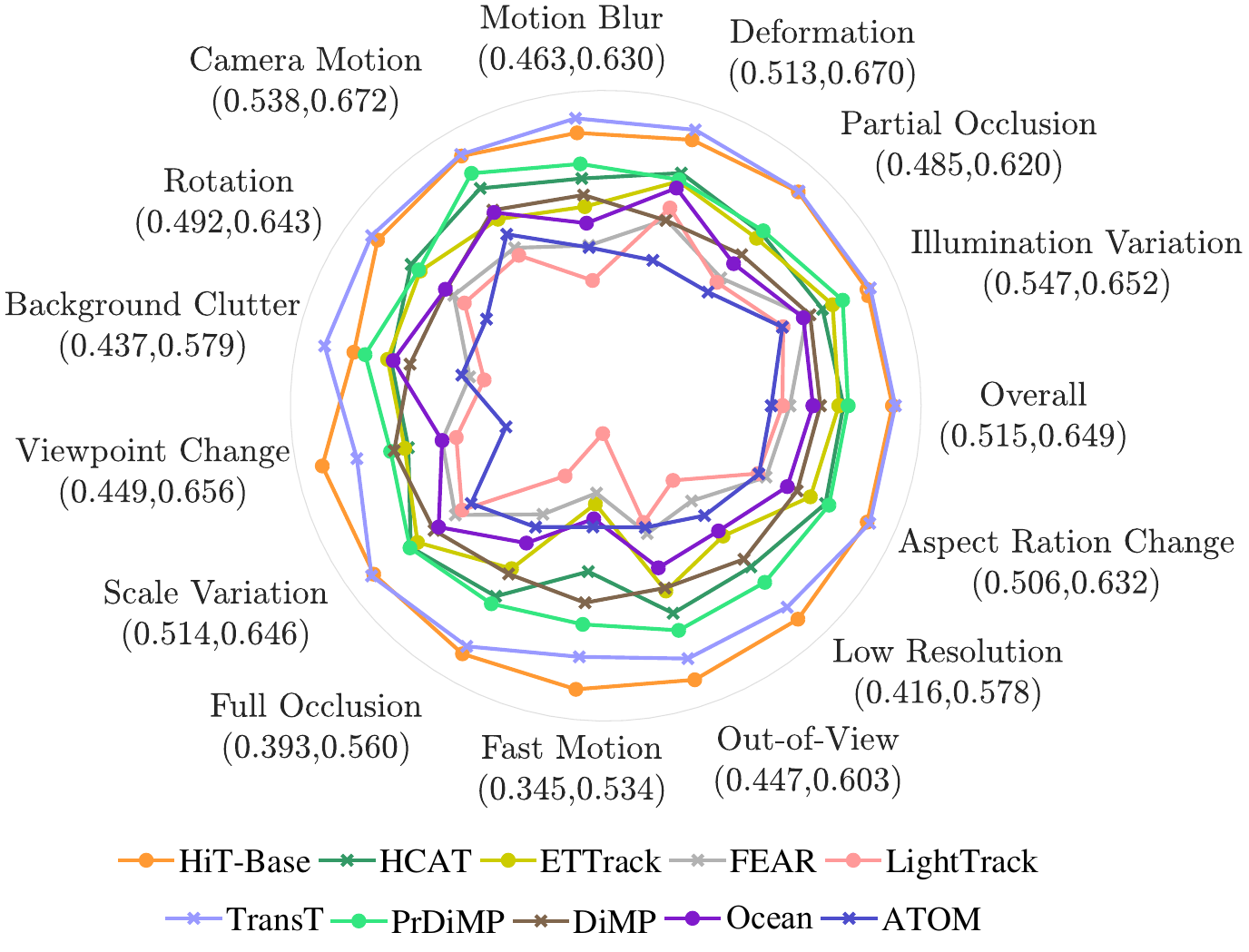}
\end{center}
   \caption{{AUC scores of different attributes on LaSOT. The numbers below each attribute represent the maximum and minimum values of that attribute among the evaluated trackers.}}
\label{fig:attributes}
\end{figure}

\noindent\textbf{TrackingNet.} TrackingNet~\cite{trackingnet} is a comprehensive dataset, encompassing diverse situations in natural scenes and featuring multiple categories. Its test set comprises 511 video sequences. As outlined in Table~\ref{tab-sota}, both HiT and DyHiT exhibit competitive performance when compared with previous real-time trackers. Notably, HiT-Base and DyHiT achieve the top two AUC scores of 80.0\% and 77.9\%, surpassing the leading real-time tracker HCAT~\cite{chen2022efficient} by 3.4\% and 1.3\%, respectively. In comparison to the non-real-time tracker STARK-ST50~\cite{Stark}, HiT-Base demonstrates comparable AUC performance (80.0 $vs$. 81.3). However, it achieves this with impressive speed gains: being $3.5 \times$ faster on the GPU, $4.7 \times$ faster on the CPU, and $4.7 \times$ faster on the AGX. This emphasizes the efficiency of HiT-Base in delivering competitive tracking results with significantly improved processing speed.

\noindent\textbf{LaSOT.} LaSOT~\cite{LaSOT} is a large-scale, long-term dataset encompassing 1400 video sequences, with 1120 training videos and 280 test videos. The performance results on LaSOT are detailed in Table~\ref{tab-sota}. HiT-Base stands out by achieving top-tier real-time results with AUC, P$_{Norm}$, and P scores of 64.6\%, 73.3\%, and 68.1\%, respectively. Additionally, DyHiT secures the second-best AUC score. In comparison with the recent efficient tracker MixformerV2-S~\cite{mixformerv2}, HiT-Base and DyHiT outperform it by 4.0\% and 1.8\%, respectively. In comparison with the non-real-time tracker TransT~\cite{TransT}, HiT-Base exhibits slightly lower performance by 0.3\% in AUC but compensates with significantly faster processing speed. Fig.~\ref{fig:attributes} shows the attribute-wise analysis of  HiT, the current state-of-the-art high-speed trackers, and some non-real-time trackers. Our HiT has completely surpassed the current state-of-the-art high-speed trackers and some non-real-time trackers (PrDiMP~\cite{PrDiMP},DiMP~\cite{DiMP}) in various attributes. Notably, HiT excels in effectively managing fast motion and viewpoint changes, showcasing superior performance in these aspects. 

\begin{table}[t]
  \centering
      \caption{{Comparison with state-of-the-art methods on additional benchmarks in AUC score.}}
  \label{tab-sota-small}
  \setlength{\tabcolsep}{1.1mm}{
    \begin{tabular}{c|lcccc}
    \toprule
    &Method&NFS&UAV123&LaSOT$_{ext}$&TNL2K\\
    \midrule[0.5pt]
    \multirow{11}*{\rotatebox{90}{Real-time}} &\cellcolor{mygray1}HiT-Base &\cellcolor{mygray1}\textbf{\textcolor{cRed}{63.6}} &\cellcolor{mygray1}\textcolor{blue}{65.6} &\cellcolor{mygray1}\textbf{\textcolor{cRed}{44.1}}&\cellcolor{mygray1}\textbf{\textcolor{cRed}{50.6}}\\
    &\cellcolor{mygray1}HiT-Small &\cellcolor{mygray1}\textcolor{cGreen}{61.8} &\cellcolor{mygray1}{63.3} &\cellcolor{mygray1}{40.4}&\cellcolor{mygray1}{47.3}\\
    &\cellcolor{mygray1}HiT-Tiny &\cellcolor{mygray1}53.2 &\cellcolor{mygray1}58.7 &\cellcolor{mygray1}35.8&\cellcolor{mygray1}41.4\\
     &\cellcolor{mygray1}DyHiT &\cellcolor{mygray1}{61.6} &\cellcolor{mygray1}\textcolor{cGreen}{64.9} &\cellcolor{mygray1}\textcolor{cGreen}{42.1}&\cellcolor{mygray1}\textcolor{blue}{49.1}\\
    &MixFormerV2-S~\cite{mixformerv2} &- &\textbf{\textcolor{cRed}{65.8}} &\textcolor{blue}{43.6} &\textcolor{cGreen}{48.3}\\
    &HCAT~\cite{chen2022efficient}&\textcolor{blue}{63.5}&62.7&{40.6}&{47.9}\\
    &FEAR~\cite{borsuk2022fear}&61.4&-&-&-\\
    &E.T.Track~\cite{ETTrack}&59.0&62.3&-&-\\
    &LightTrack~\cite{yan2021lighttrack}&55.3&62.5&-&-\\
    &ATOM~\cite{ATOM}&58.4&{64.2}&37.6&39.2\\ 
    &ECO~\cite{danelljan2017eco}&46.6&53.2&22.0&31.7\\
    \midrule[0.5pt]
    \multirow{12}*{\rotatebox{90}{Non-real-time}}
    &MCITrack-B224~\cite{mcitrack} &70.6 &70.5 &\emph{\uline{54.6}} &62.9\\
    &SUTrack-B224~\cite{sutrack} &\emph{\uline{71.3}} &\emph{\uline{71.7}} &53.1 &\emph{\uline{65.0}}\\
    &MixFormerV2-B~\cite{mixformerv2} &- &69.9 &50.6 &57.4\\
    &SeqTrack-B256~\cite{SeqTrack} &67.6 &69.2 &49.5 &54.9\\
    &ARTrack-256~\cite{artrack} &64.3 &67.7 &46.4 &57.5 \\
    &OSTrack-256~\cite{ostrack}&64.7&68.3&47.4&55.9\\
    &TransT~\cite{TransT}&65.7&{69.1}&44.4&50.7\\
    &TrDiMP~\cite{wang2021transformer}&66.5&67.5&-&-\\ 
    &TrSiam~\cite{wang2021transformer}&65.8&67.4&-&-\\
    &PrDiMP~\cite{PrDiMP}&63.5&68.0&-&45.9\\ 
    &DiMP~\cite{DiMP} &62.0&65.3&39.2&44.7\\
    &SiamRPN++~\cite{SiamRPNplusplus}&50.2&61.6&34.0&41.2\\
    \bottomrule
    \end{tabular}
  }
\end{table}

\noindent\textbf{GOT-10k.} GOT-10k~\cite{GOT10K} is a large-scale and challenging dataset, comprising 10,000 training sequences and 180 test sequences. The tracking results, detailed in Table~\ref{tab-sota}, showcase HiT-Base achieving the second-best real-time performance with an AO score of 64.0\%. Simultaneously, DyHiT secures the third-best AO score, reaching 62.9\%. Notably, HiT-Base outperforms the recent efficient tracker FEAR~\cite{borsuk2022fear} and MixformerV2-S~\cite{mixformerv2} by a significant margin of 2.1\% and 1.9\%, respectively. 

\noindent\textbf{NFS.}  NFS~\cite{NFS} is a challenging dataset renowned for its fast-moving objects, comprising 100 video sequences. As depicted in Table~\ref{tab-sota-small}, HiT-Base and HiT-Small attain the top and third positions, respectively, in terms of real-time performance. Notably, HiT-Base outperforms FEAR by a margin of 2.2\%.

\noindent\textbf{UAV123.} UAV123~\cite{UAV} is specifically designed for low-altitude UAVs, comprising 123 video clips. As shown  in Table~\ref{tab-sota-small}, HiT-Base and DyHiT attain the second and third positions among real-time trackers, with AUC scores of 65.6\% and 64.9\%, respectively.

\noindent\textbf{LaSOT$_{ext}$.} LaSOT$_{ext}$~\cite{lasot_journal} is a recently introduced tracking dataset comprising 150 videos, serving as an extension to LaSOT. The performance of HiT and DyHiT on LaSOT$_{ext}$ are detailed in Table~\ref{tab-sota-small}. HiT-Base, HiT-Small, and DyHiT exhibit competitive results with AUC scores of 44.1\%, 40.4\%, and 42.1\%, respectively. In comparison to the non-real-time tracker TransT~\cite{TransT}, HiT-Base demonstrates only a 0.3\% decrease in performance while operating at a remarkable speed of 4.7 $\times$ faster on AGX.

\noindent\textbf{TNL2K.} TNL2K~\cite{TNL2K} is a comprehensive benchmark for tracking, featuring 1300 training sequences and 700 test sequences. 
As shown in Table~\ref{tab-sota-small}, HiT-Base and DyHiT secure the top two positions for real-time performance, surpassing the leading real-time tracker MixFormerV2-S~\cite{mixformerv2} by 2.3\% and 0.8\%, respectively. Compared to the non-real-time tracker TransT~\cite{TransT}, HiT-Base demonstrates nearly identical AUC performance (50.6\% vs. 50.7\%). However, HiT-Base achieves this with significant speed advantages: it is $2.8 \times$ faster on GPU, $6.6 \times$ faster on CPU, and $4.7 \times$ faster on AGX.

\begin{table}[t]
\centering
\caption{{Real-time experiment on VOT2021. The evaluation is conducted using box evaluation. For real-time trackers, the results are obtained on the Nvidia Jetson AGX, while for non-real-time trackers, the evaluation is performed on an 2080 Ti GPU.}}
\label{tab-vot}
  \setlength{\tabcolsep}{1.7mm}{
    \begin{tabular}{c|lccc}
    \toprule[0.5pt]
    &Method &EAO&Accuracy&Robustness\\
    \midrule[0.5pt]
    \multirow{11}*{\rotatebox{90}{Real-time}}
    &\cellcolor{mygray1}HiT-Base &\cellcolor{mygray1}\textcolor{blue}{0.252}&\cellcolor{mygray1}\textcolor{cRed}{\textbf{0.447}}&\cellcolor{mygray1}0.688 \\
    &\cellcolor{mygray1}DyHiT$_{1}$ &\cellcolor{mygray1}\textcolor{blue}{0.252}&\cellcolor{mygray1}\textcolor{cRed}{\textbf{0.447}}&\cellcolor{mygray1}0.688\\
    &\cellcolor{mygray1}DyHiT$_{0.75}$ &\cellcolor{mygray1}\textcolor{cRed}{\textbf{0.253}}&\cellcolor{mygray1}\textcolor{blue}{0.446}&\cellcolor{mygray1}\textcolor{cGreen}{0.693}\\
    &\cellcolor{mygray1}DyHiT$_{0.65}$ &\cellcolor{mygray1}\textcolor{blue}{0.252}&\cellcolor{mygray1}\textcolor{cGreen}{0.442}&\cellcolor{mygray1}\textcolor{blue}{0.697}\\
    &\cellcolor{mygray1}DyHiT$_{0}$ &\cellcolor{mygray1}\textcolor{cGreen}{0.250}&\cellcolor{mygray1}0.439&\cellcolor{mygray1}0.651\\
    &MixformerV2-S\cite{mixformerv2} &\textcolor{blue}{0.252}&0.441&\textcolor{cRed}{\textbf{0.702}}\\
    &FEAR ~\cite{borsuk2022fear} &\textcolor{cGreen}{0.250} &{0.436} &{0.655} \\
    &STARK-S~\cite{Stark} &{0.237} &{0.407} &0.631 \\
    &STARK-Lightning~\cite{Stark} &{0.204} &{0.391} &0.565\\
    &LightTrack~\cite{yan2021lighttrack} &0.225 &0.391 &{0.641} \\
    &E.T.Track~\cite{ETTrack} &0.224&0.372&0.631\\
    \midrule[0.1pt]
    \multirow{6}*{\rotatebox{90}{Non-real-time}}
    &MCITrack-B224~\cite{mcitrack} &\emph{\uline{0.361}}&0.463&\emph{\uline{0.889}}\\
    &SUTrack-B224~\cite{sutrack} &0.339&0.469&0.859\\
    &ARTrack-256~\cite{artrack} &0.318&0.472&0.814\\
    &MixFormerV2-B~\cite{mixformerv2} &0.314&\emph{\uline{0.475}}&0.812\\
    &SeqTrack-B256~\cite{SeqTrack} &0.309&0.471&0.799\\
    &OSTrack~\cite{ostrack} &0.296&0.457&0.777\\
    &TransT-M~\cite{transtm} &0.315&0.463&0.823\\
    &DualTFRst~\cite{dualrt}  &0.327&0.470&0.829\\
    &STARK-RT~\cite{Stark} &0.280&0.467&0.745 \\
    \bottomrule[0.5pt]
    \end{tabular}
  }
\end{table}

\begin{table}[t]
\centering
\caption{{Speed comparison under low-power mode (10W) on Nvidia Jetson AGX.}}
\label{tab-agx-10w}
  \setlength{\tabcolsep}{0.4mm}{
    \begin{tabular}{c|ccc|cc}
    \toprule[0.5pt]
    Model&DyHiT &HiT-Base&HiT-Small&FEAR&MixFormerV2-S\\
    \midrule[0.5pt]
    {{Speed(fps)}}&32&16&28&10&18\\
    \bottomrule[0.5pt]
    \end{tabular}
  }
\end{table}

\noindent\textbf{VOT.}  VOT competition is regarded as one of the most challenging competitions in visual tracking.  We also conduct VOT real-time experiments on VOT2021 benchmark~\cite{vot2021}. The results are shown in the Table~\ref{tab-vot}. 
{In DyHiT, the subscript indicates the scene-splitting threshold: when the predicted score is higher than this threshold, the scene is classified as easy; otherwise, it is considered challenging. When the subscript is set to 1, only Route2(full pipeline) is used; when set to 0, only Route1 is used. As shown, DyHiT$_{0.75}$ achieves the best real-time performance with an EAO score of 0.253. HiT-Base,  DyHiT$_{1}$, and  DyHiT$_{0.65}$ achieve the second-best real-time performance with an EAO score of 0.252.  DyHiT$_{0}$ achieves the third-best result with an EAO score of 0.250.}

\noindent\textbf{Analysis of VOT Real-Time Experiments.}
{
In  VOT , the real-time threshold is defined as 20 fps. When a model fails to reach this threshold, its performance becomes significantly constrained by speed. However, once the model's speed exceeds 20 fps, its tracking performance is determined solely by its inherent capabilities—faster speed does not necessarily translate to better performance.
The results from the VOT real-time experiments reveal a notable performance gap between real-time and non-real-time trackers. This disparity arises from the experimental setup: to simulate real-world deployment scenarios, real-time trackers were evaluated on the Nvidia Jetson AGX platform. Due to their large computational requirements and slower inference speeds, non-real-time trackers cannot run effectively on AGX. As a result, we conducted their evaluations on a desktop-class Nvidia 2080Ti GPU. Since most non-real-time trackers achieve frame rates exceeding 20 fps on the 2080Ti, their performance is not limited by speed. Moreover, their larger model sizes and higher parameter counts generally lead to better performance.
However, in practical applications, tracking algorithms are often deployed on low-power, resource-constrained devices where non-real-time trackers are not feasible. In contrast, real-time trackers can be effectively deployed in such environments. Additionally, in real-world perception systems, tracking is typically integrated with other tasks such as object detection and segmentation. This means the computing resources allocated to the tracker are further limited. Therefore, achieving real-time performance under low computational budgets becomes a critical advantage.
Our proposed DyHiT and HiT models are specifically designed for such scenarios. To further validate their practical utility, we conducted additional experiments under a more constrained setting by configuring the Jetson AGX to low-power mode (10W power limit). As shown in Table~\ref{tab-agx-10w}, only DyHiT and HiT-Small were able to maintain frame rates above the 20 fps threshold in this setting. This result underscores the high practical value of our models for real-time visual tracking in low-power, real-world deployments.}

\begin{table*}[t]\small
  \centering
  \caption{Evaluation of DyTrackers on TrackingNet~\cite{trackingnet}, LaSOT~\cite{LaSOT}, and GOT-10k~\cite{GOT10K} benchmarks.  $\Delta_A$ denotes the performance (AUC or AO) change (averaged over benchmarks) compared with the baseline. $\Delta_S$ denotes the speed change compared withe the baseline. The results on GOT-10k are trained with additional datasets. We tested on the Nvidia GeForce RTX 2080 Ti GPU using a single thread on the GOT-10k test set. Subsequently, we submitted the test results to the GOT-10k evaluation server~\cite{GOT10K} for evaluation, obtaining the speed.}
  \label{tab-dysota}
  \setlength{\tabcolsep}{0.9mm}{
  \begin{tabular}{l|c ccc c ccc c ccc c c c c}
    \toprule
    \multirow{2}*{Method} &  \multicolumn{3}{c}{TrackingNet}&& \multicolumn{3}{c}{LaSOT} &&  \multicolumn{3}{c}{GOT-10k} &&\multirow{2}*{Speed (\emph{fps})} &\multirow{2}*{$\Delta_A$} &\multirow{2}*{$\Delta_S$ (\emph{fps})} \\
\cmidrule(r){2-4}
\cmidrule(r){6-8}
\cmidrule(r){10-12}
   & AUC&P$_{Norm}$&P && AUC&P$_{Norm}$&P  && AO&SR$_{0.5}$&SR$_{0.75}$\\        \midrule[0.5pt]
    \cellcolor{mygray1}OSTrack-256~\cite{ostrack}&\cellcolor{mygray1}83.1 &\cellcolor{mygray1}87.8 &\cellcolor{mygray1}82.0 &\cellcolor{mygray1}&\cellcolor{mygray1}69.1 &\cellcolor{mygray1}78.7 &\cellcolor{mygray1}75.2 &\cellcolor{mygray1}&\cellcolor{mygray1}73.7  &\cellcolor{mygray1}83.0 &\cellcolor{mygray1}71.0  &\cellcolor{mygray1}&\cellcolor{mygray1}70 &\cellcolor{mygray1}- &\cellcolor{mygray1}- \\
    \cellcolor{mygray2}DyOSTrack-256&\cellcolor{mygray2}82.5 &\cellcolor{mygray2}87.0 &\cellcolor{mygray2}80.6 &\cellcolor{mygray2}&\cellcolor{mygray2}69.5 &\cellcolor{mygray2}79.1 &\cellcolor{mygray2}74.6 &\cellcolor{mygray2}&\cellcolor{mygray2}73.7 &\cellcolor{mygray2}83.5 &\cellcolor{mygray2} 71.2 &\cellcolor{mygray2}&\cellcolor{mygray2}110 &\cellcolor{mygray2}\textbf{-0.06} &\cellcolor{mygray2}\bm{$1.57 \times$} \\
    \cellcolor{brown10}SeqTrack-B256~\cite{SeqTrack}&\cellcolor{brown10}83.3 &\cellcolor{brown10}88.3 &\cellcolor{brown10}82.2 &\cellcolor{brown10}&\cellcolor{brown10}69.9 &\cellcolor{brown10}79.7 &\cellcolor{brown10}76.3 &\cellcolor{brown10}&\cellcolor{brown10}76.1 &\cellcolor{brown10}85.5 &\cellcolor{brown10}74.7  &\cellcolor{brown10}&3\cellcolor{brown10}1 &\cellcolor{brown10}- &\cellcolor{brown10}- \\
    \cellcolor{brown20}DySeqTrack-B256&\cellcolor{brown20}82.6 &\cellcolor{brown20}87.6 &\cellcolor{brown20}80.7 &\cellcolor{brown20}&\cellcolor{brown20}69.9 &\cellcolor{brown20}80.1 &\cellcolor{brown20}75.2 &\cellcolor{brown20}&\cellcolor{brown20}75.8  &\cellcolor{brown20}86.1 &\cellcolor{brown20}74.0 &\cellcolor{brown20}&\cellcolor{brown20}83 &\cellcolor{brown20}\textbf{-0.33} &\cellcolor{brown20}\bm{$2.68 \times$} \\
    \cellcolor{mygray1}MixFormer-22k~\cite{mixformer}&\cellcolor{mygray1}83.1&\cellcolor{mygray1}88.1&\cellcolor{mygray1}81.6 &\cellcolor{mygray1}&\cellcolor{mygray1}69.2 &\cellcolor{mygray1}78.7 &\cellcolor{mygray1}74.7 &\cellcolor{mygray1}&\cellcolor{mygray1}72.5  &\cellcolor{mygray1}81.6 &\cellcolor{mygray1}70.1  &\cellcolor{mygray1}&\cellcolor{mygray1}32 &\cellcolor{mygray1}- &\cellcolor{mygray1}- \\
    \cellcolor{mygray2}DyMixFormer-22k&\cellcolor{mygray2}82.8 &\cellcolor{mygray2}87.6 &\cellcolor{mygray2}80.8 &\cellcolor{mygray2}&\cellcolor{mygray2}69.5 &\cellcolor{mygray2}79.1 &\cellcolor{mygray2}74.2 &\cellcolor{mygray2}&\cellcolor{mygray2}72.7  &\cellcolor{mygray2}82.1 &\cellcolor{mygray2}70.0  &\cellcolor{mygray2}&\cellcolor{mygray2}70 &\cellcolor{mygray2}\textbf{+0.07} &\cellcolor{mygray2}\bm{$2.19 \times$} \\
    \cellcolor{brown10}Sim-B/16~\cite{simtrack}&\cellcolor{brown10}82.3 &\cellcolor{brown10}86.5 &\cellcolor{brown10}- &\cellcolor{brown10}&\cellcolor{brown10}69.3 &\cellcolor{brown10}78.5 &\cellcolor{brown10}74.0 &\cellcolor{brown10}&\cellcolor{brown10}70.6  &\cellcolor{brown10}80.0 &\cellcolor{brown10}67.7  &\cellcolor{brown10}&\cellcolor{brown10}89 &\cellcolor{brown10}- &\cellcolor{brown10}- \\
    \cellcolor{brown20}DySim-B/16&\cellcolor{brown20}81.9 &\cellcolor{brown20}86.3 &\cellcolor{brown20}79.3 &\cellcolor{brown20}&\cellcolor{brown20}69.1 &\cellcolor{brown20}78.6 &\cellcolor{brown20}73.6 &\cellcolor{brown20}&\cellcolor{brown20}70.7  &\cellcolor{brown20}80.3 &\cellcolor{brown20}67.5  &\cellcolor{brown20}&\cellcolor{brown20}108 &\cellcolor{brown20}\textbf{-0.17} &\cellcolor{brown20}\bm{$1.21 \times$} \\
    \cellcolor{mygray1}STARK-S50~\cite{Stark}&\cellcolor{mygray1}80.3 &\cellcolor{mygray1}85.1 &\cellcolor{mygray1}- &\cellcolor{mygray1}&\cellcolor{mygray1}65.8 &\cellcolor{mygray1}75.2 &\cellcolor{mygray1}69.7 &\cellcolor{mygray1}&\cellcolor{mygray1}68.9  &\cellcolor{mygray1}78.7 &\cellcolor{mygray1}64.2  &\cellcolor{mygray1}&\cellcolor{mygray1}46 &\cellcolor{mygray1}- &\cellcolor{mygray1}- \\
    \cellcolor{mygray2}DySTARK-S50&\cellcolor{mygray2}80.0 &\cellcolor{mygray2}84.5 &\cellcolor{mygray2}76.8 &\cellcolor{mygray2}&\cellcolor{mygray2}66.7 &\cellcolor{mygray2}76.1 &\cellcolor{mygray2}70.2 &\cellcolor{mygray2}&\cellcolor{mygray2}68.8  &\cellcolor{mygray2}78.9 &\cellcolor{mygray2}63.9  &\cellcolor{mygray2}&\cellcolor{mygray2}89 &\cellcolor{mygray2} \textbf{+0.17}&\cellcolor{mygray2}\bm{$1.93 \times$} \\
    \cellcolor{brown10}STARK-ST101~\cite{Stark}&\cellcolor{brown10}82.0 &\cellcolor{brown10}86.9 &\cellcolor{brown10}- &\cellcolor{brown10}&\cellcolor{brown10}67.1 &\cellcolor{brown10}76.9 &\cellcolor{brown10}72.2 &\cellcolor{brown10}&\cellcolor{brown10}70.3  &\cellcolor{brown10}80.2 &\cellcolor{brown10}65.9  &\cellcolor{brown10}&\cellcolor{brown10}28 &\cellcolor{brown10}- &\cellcolor{brown10}- \\
    \cellcolor{brown20}DySTARK-ST101&\cellcolor{brown20}81.7 &\cellcolor{brown20}86.6 &\cellcolor{brown20}78.7 &\cellcolor{brown20}&\cellcolor{brown20}67.7 &\cellcolor{brown20}77.3 &\cellcolor{brown20}72.2 &\cellcolor{brown20}&\cellcolor{brown20}70.1  &\cellcolor{brown20}80.3 &\cellcolor{brown20}65.7  &\cellcolor{brown20}&\cellcolor{brown20}82 &\cellcolor{brown20}\textbf{+0.03} &\cellcolor{brown20}\bm{$2.93 \times$} \\
    \cellcolor{mygray1}TransT~\cite{TransT}&\cellcolor{mygray1}81.4 &\cellcolor{mygray1}86.7 &\cellcolor{mygray1}80.3 &\cellcolor{mygray1}&\cellcolor{mygray1}64.9 &\cellcolor{mygray1}73.8 &\cellcolor{mygray1}\cellcolor{mygray1}69.0 &\cellcolor{mygray1}&\cellcolor{mygray1}72.3  &\cellcolor{mygray1}82.3 &\cellcolor{mygray1}68.2  &\cellcolor{mygray1}&\cellcolor{mygray1}58 &\cellcolor{mygray1}- &\cellcolor{mygray1}- \\
    \cellcolor{mygray2}DyTransT&\cellcolor{mygray2}81.0 &\cellcolor{mygray2}86.3 &\cellcolor{mygray2}79.0 &\cellcolor{mygray2}&\cellcolor{mygray2}64.9 &\cellcolor{mygray2}74.3 &\cellcolor{mygray2}68.9 &\cellcolor{mygray2}&\cellcolor{mygray2}71.2 &\cellcolor{mygray2}81.5 &\cellcolor{mygray2}\cellcolor{mygray2}66.7 &\cellcolor{mygray2}&\cellcolor{mygray2}108 &\cellcolor{mygray2} \textbf{-0.50}&\cellcolor{mygray2}\bm{$1.86 \times$} \\ \bottomrule 
\end{tabular}}
\end{table*}

\subsection{Evaluation of DyTrackers}
Our acceleration method exhibits excellent generalization ability, allowing it to seamlessly integrate with various high-performance trackers to create different DyTrackers. We apply our acceleration scheme to various high-performance trackers, including OSTrack-256~\cite{ostrack}, SeqTrack-256~\cite{SeqTrack}, MixFormer-22k~\cite{mixformer}, Sim-B/16~\cite{simtrack}, STARK-S50~\cite{Stark}, STARK-ST101~\cite{Stark}, and TransT~\cite{TransT}, resulting in a series of DyTrackers, namely DyOSTrack-256, DySeqTrack-B256, DyMixFormer-22k, DySim-B/16, DySTARK-S50, DySTARK-ST101, and DyTransT.
We conduct a comprehensive comparison of DyTrackers with their base trackers across seven datasets, evaluating both speed and accuracy.
The evaluation results are presented in Tables~\ref{tab-dysota} and~\ref{tab-dyhit-small}. 
\begin{table}[t]
  \centering
      \caption{Evaluation of DyTrackers on additional benchmarks in AUC score. $\Delta_A$ denotes the performance (AUC) change (averaged over benchmarks) compared to the corresponding base tracker.}
  \label{tab-dyhit-small}
  \setlength{\tabcolsep}{0.5mm}{
    \begin{tabular}{l|ccccc}
    \toprule
    Method&NFS&UAV123&LaSOT$_{ext}$&TNL2K&$\Delta_A$\\
    \midrule[0.5pt]
\cellcolor{mygray1}OSTrack-256~\cite{ostrack}&\cellcolor{mygray1}64.7&\cellcolor{mygray1}68.3 &\cellcolor{mygray1}47.4 &\cellcolor{mygray1}54.3 &\cellcolor{mygray1}- \\
    \cellcolor{mygray2}DyOSTrack-256&\cellcolor{mygray2}65.8 &\cellcolor{mygray2}70.8 &\cellcolor{mygray2}46.7 &\cellcolor{mygray2}55.5 &\cellcolor{mygray2}\textbf{+1.03} \\
    \cellcolor{brown10}SeqTrack-B256~\cite{SeqTrack}&\cellcolor{brown10}67.6 &\cellcolor{brown10}69.2 &\cellcolor{brown10}49.5 &\cellcolor{brown10}54.9 &\cellcolor{brown10}-\\
    \cellcolor{brown20}DySeqTrack-B256&\cellcolor{brown20}66.9 &\cellcolor{brown20}69.7 &\cellcolor{brown20}49.0 &\cellcolor{brown20}56.4 &\cellcolor{brown20}\textbf{+0.20}  \\
    \cellcolor{mygray1}MixFormer-22k~\cite{mixformer}&\cellcolor{mygray1}66.4 &\cellcolor{mygray1}70.4 &\cellcolor{mygray1}49.8 &\cellcolor{mygray1}55.4  &\cellcolor{mygray1}-\\
    \cellcolor{mygray2}DyMixFormer-22k&\cellcolor{mygray2}66.9 &\cellcolor{mygray2}71.4 &\cellcolor{mygray2}49.9 &\cellcolor{mygray2}55.5 &\cellcolor{mygray2}\textbf{+0.43} \\
    \cellcolor{brown10}Sim-B/16~\cite{simtrack}&\cellcolor{brown10}64.5 &\cellcolor{brown10}69.8 &\cellcolor{brown10}48.8 &\cellcolor{brown10}54.8 &\cellcolor{brown10}-\\
    \cellcolor{brown20}DySim-B/16&\cellcolor{brown20}64.6 &\cellcolor{brown20}70.1 &\cellcolor{brown20}48.9 &\cellcolor{brown20}54.5 &\cellcolor{brown20} \textbf{+0.05}\\
    \cellcolor{mygray1}STARK-S50~\cite{Stark}&\cellcolor{mygray1}66.0 &\cellcolor{mygray1}68.8 &\cellcolor{mygray1}46.6 &\cellcolor{mygray1}53.2 &\cellcolor{mygray1}-\\
    \cellcolor{mygray2}DySTARK-S50&\cellcolor{mygray2}66.3 &\cellcolor{mygray2}69.7 &\cellcolor{mygray2}46.3 &\cellcolor{mygray2}53.1 &\cellcolor{mygray2}\textbf{+0.20} \\
    \cellcolor{brown10}STARK-ST101~\cite{Stark}&\cellcolor{brown10}67.7 &\cellcolor{brown10}67.7 &\cellcolor{brown10}46.6 &\cellcolor{brown10}55.1 &\cellcolor{brown10}-\\
    \cellcolor{brown20}DySTARK-ST101&\cellcolor{brown20}67.3 &\cellcolor{brown20}68.6 &\cellcolor{brown20}46.5 &\cellcolor{brown20}55.2 &\cellcolor{brown20}\textbf{+0.13}  \\
    \cellcolor{mygray1}TransT~\cite{TransT}&\cellcolor{mygray1}65.7 &\cellcolor{mygray1}69.1 &\cellcolor{mygray1}44.4 &\cellcolor{mygray1}50.7 &\cellcolor{mygray1}-\\
    \cellcolor{mygray2}DyTransT&\cellcolor{mygray2}65.4 &\cellcolor{mygray2}67.7 &\cellcolor{mygray2}44.3 &\cellcolor{mygray2}53.5 &\cellcolor{mygray2}\textbf{+0.25}\\
    \bottomrule
    \end{tabular}
  }
\end{table}
It can be observed that after applying our acceleration method, there is a significant improvement in the speed of these high-performance trackers. 
For instance, the well-known one-stream tracker OSTrack-256 achieves a speed of 110 fps, which is $1.57 \times$ faster than before. Moreover, there is an improvement in accuracy, reaching 69.5\% AUC on LaSOT, which is 0.4\% higher than before, 65.8\% AUC on NFS, an increase of 1.1\%, 70.8\% AUC on UAV123, a gain of 2.5\%, and 55.5\% AUC on TNL2K, a rise of 1.2\%. 
In summary, these mainstream high-performance trackers, when accelerated using our method, experience significant speed improvements with almost no loss in accuracy. This demonstrates the effectiveness and generalization ability of our acceleration method.

\subsection{Ablation and Analysis}
In this section, we present a series of ablation experiments to thoroughly analyze our approach. These experiments primarily include: ablation analysis of the Bridge Module, ablation analysis of the dual-image position encoding, a study on the generalization of the HiT framework, an investigation of the speed-accuracy trade-off in DyHiT, ablation analysis of the efficient dynamic routing structure in DyHiT, and a discussion on the training methods of DyHiT.
It is essential to highlight that, for the ablation study concerning HiT, we employ HiT-Base as the baseline model. All HiT models in the ablation experiments are trained for 500 epochs.

\begin{figure}[t]
\begin{center}
\includegraphics[width=1\linewidth]{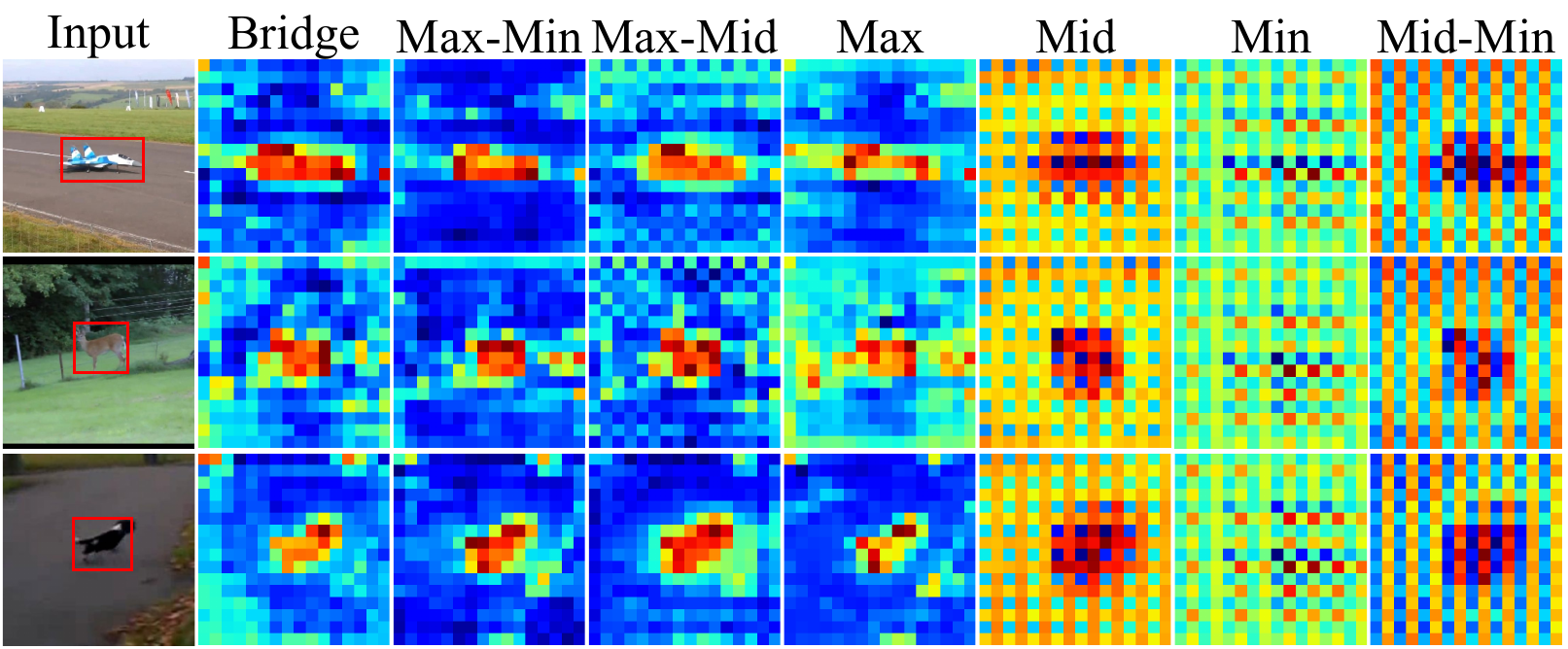}
\end{center}
   \caption{Visualization of the attention maps in the corner head of different feature combining manners. Bridge means our default manner, Max-Min means combining the Max and the Min features, Max-Mid means combining the Max and the Mid features, Max, Mid, and Min mean only using the Max feature, Mid feature, and Min feature, respectively.}
\label{fig:attn_vis}
\end{figure}

\noindent\textbf{Different combinations of features.} 
To verify the effectiveness of the Bridge Module and explore the importance of different features, we compare various feature combinations within the Bridge Module. Table~\ref{tab-components} presents the results, where Max, Mid, and Min denote the features of the first, second, and third stages of the transformer, respectively. To ensure a fair comparison, the features are upsampled to the same resolution. The first row (\#1) represents our default setting. Initially, without utilizing the Bridge Module, we make predictions based on independent Max, Mid, and Min features. Table~\ref{tab-components} (\#2, \#3, and \#4) indicates that these methods result in inferior performance, underscoring the effectiveness of feature fusion achieved by our Bridge Module. Subsequently, Table~\ref{tab-components} (\#5, \#6, and \#7) presents the results of alternative combinations, and our default method stands out as the most effective. In our default approach, incorporating all three features provides a richer blend of semantic and detailed information, contributing to superior results.
To gain a deeper understanding of the Bridge Module, we visualize the attention map in the corner head for different feature combination methods, as depicted in Fig.~\ref{fig:attn_vis}. In the visualization results, we make two key observations: 
(1) Collapse Phenomenon: Methods that exclude the Max feature exhibit a collapse phenomenon. Taking the Mid method as an example, the final feature originates from the second stage of the transformer and is up-sampled by a factor of 2. Consequently, one pixel on the feature map is up-sampled to four pixel points. In the visualization result, we observe that the attention collapses to a relatively fixed distribution for every four upsampling grids. The Min column and the Mid-Min column show similar behavior to the Mid column. This highlights that even if the deep feature is up-sampled to a larger resolution, it does not inherently provide more detailed information. Therefore, the involvement of shallow, large-resolution features becomes crucial for supplementing information.
(2) Improved Accuracy with Min Feature: The attention map of our default method, which includes all three features, demonstrates higher accuracy compared to methods that exclude the Min feature. This finding underscores that leveraging deep features to complement semantic information significantly enhances the discriminative ability of the model.

\begin{table}[t]\scriptsize
  \centering
     \caption{
     {Ablation study on different feature combination methods and the use of Shrink Attention in terms of AUC performance. The default setting is shown in \textcolor{gray}{gray}. The best results are highlighted in \textbf{\textcolor{cRed}{red}}. Max, Mid, and Min denote the features from the first, second, and third stages of the transformer, respectively. SA denotes Shrink Attention.
     }}
  \label{tab-components}
  \setlength{\tabcolsep}{1.1mm}{
    \begin{tabular}{c|cccc|ccc}
    \toprule
    \#&Max&Mid&Min&SA&LaSOT&TrackingNet&GOT-10k\\
    \midrule[0.5pt] 
    \rowcolor{mygray1}
    1&\checkmark&\checkmark&\checkmark&\checkmark&\textbf{\textcolor{cRed}{63.7}}&\textbf{\textcolor{cRed}{78.9}}&\textbf{\textcolor{cRed}{65.4}}\\
    2&\checkmark&&&\checkmark&62.1&78.3&63.4\\
    3&&\checkmark&&\checkmark&61.9&78.1&64.1\\
    4&&&\checkmark&\checkmark&57.9&73.0&62.1\\
    5&\checkmark&\checkmark&&\checkmark&62.6&78.7&63.0\\
    6&\checkmark&&\checkmark&\checkmark&58.8&77.2&60.4\\
    7&&\checkmark&\checkmark&\checkmark&60.3&78.4&63.6\\
    8&\checkmark&\checkmark&\checkmark&&58.1&76.1&59.2\\
    \bottomrule
    \end{tabular}
  }
\end{table}

{\noindent{\textbf{Shrink Attention.} In our HiT-Base model, we follow LeViT~\cite{graham2021levit} and use the Shrink Attention to downsample the feature map. Here, we compare different downsample methods. As shown in Table \ref{tab-components}. 
The eighth row (\#8)  indicates that we do not use Shrink Attention and instead downsample the feature map using a convolution with a stride of 2. As a result, a significant performance drop is observed, with decreases of 5.6\%, 2.8\%, and 6.2\% on LaSOT, TrackingNet, and GOT-10k, respectively.}}

\begin{table}[t]
\centering
\caption{Comparison of different Position Encoding (PE) in AUC score. DI denotes our dual-image PE. Abs denotes the absolute PE. Sep denotes the relative PE which encodes the template and search region separately. Ver and Hor denote the joint encoding of the template and search images in a vertical and horizontal arrangement, respectively.
}
\label{ablation on position encoding}
\setlength{\tabcolsep}{3.8mm}{
\begin{tabular}{l|l|cccc}
\toprule
\#&PE &LaSOT	&TrackingNet	&GOT-10k\\
\midrule[0.5pt]
\rowcolor{mygray1}
1&DI  &\textbf{\textcolor{cRed}{63.7}}	 &\textbf{\textcolor{cRed}{78.9}} &\textbf{\textcolor{cRed}{65.4}}\\
2&Abs &60.2 &77.2 &61.2\\
3&Sep &62.4 &77.6 &63.1\\
4&Ver &61.1 &78.4 &63.5\\
5&Hor &61.0 &78.5 &63.7\\
\bottomrule
\end{tabular}}
\end{table}

\noindent\textbf{Different Position Encoding.} In previous transformer-based trackers~\cite{TransT, Stark}, the position information of the search image and the template image is encoded separately. In our dual-image position encoding method, we assign a unique position for each image and jointly encode their position information. Here, we compare our method with four potential encoding methods, and the results are reported in Table~\ref{ablation on position encoding}. First, we compare our method with absolute position encoding (denoted as Abs) and relative position encoding, which encodes the search and template images separately (denoted as Sep). Table~\ref{ablation on position encoding} (\#1, \#2 and \#3) indicates that these methods show inferior performance to our dual-image position encoding. The separate encoding fails to model the positional relationship between the search and template images, introducing overlapping positions and leading to inferior performance. Second, in our dual-image position encoding, we explore different arrangements of the template and search regions. By default, we diagonally arrange the template and the search region, as shown in Fig.~\ref{fig:PE-our}. Here we compare it with two other arrangements: the vertical arrangement (denoted as Ver) and the horizontal arrangement (denoted as Hor). Table~\ref{ablation on position encoding} (\#1, \#4, and \#5) shows that the default diagonal arrangement achieves the best performance. In the vertical and horizontal arrangements, the horizontal and vertical positions of the template and the search region overlap, leading to information loss. The diagonal arrangement assigns unique horizontal and vertical positions for the template and the search region, providing more informative encoding. Therefore, we choose the diagonal arrangement.

\begin{table}[t]
\centering
\caption{HiT with different lightweight hierarchical vision transformers.}
\label{different backbone}
\setlength{\tabcolsep}{0.4mm}{
\begin{tabular}{c|c|cc}
\toprule
& &LeViT-384~\cite{graham2021levit} &PVT-Small~\cite{wang2021pyramid} \\
\midrule[0.5pt]
\multirow{3}{*}{Benchmarks}&LaSOT&63.7&63.9\\
&TrackingNet&78.9&78.4\\
&GOT-10k&65.4&64.8\\
\midrule[0.5pt]
\multirow{3}{*}{PyTorch Speed (\emph{fps})}&GPU&175&91\\
&CPU&33&22\\
&AGX&61&30\\
\midrule[0.5pt]
\multirow{3}{*}{ONNX Speed (\emph{fps})}&GPU&274&133\\
&CPU&42&25\\
&AGX&75&32\\
\bottomrule
\end{tabular}}
\end{table}

\noindent\textbf{Different Backbones.} 
To assess the generalization of our HiT framework, we extend our architecture with another hierarchical vision transformer, PVT~\cite{wang2021pyramid}. The results are presented in Table~\ref{different backbone}. We utilize PVT-Small~\cite{wang2021pyramid} as the transformer backbone, keeping the other components consistent with HiT-Base.
As shown in Table~\ref{different backbone}, HiT with PVT-Small achieves 63.9\% AUC on LaSOT, 78.4\% AUC on TrackingNet, and 64.8\% AO on GOT-10k, while maintaining real-time speeds on all three platforms. This result is competitive compared to our base model with LeViT-384 and other efficient trackers, showcasing the superior generalization ability of our framework.

\begin{figure}[t]
\begin{center}
\includegraphics[width=1\linewidth]{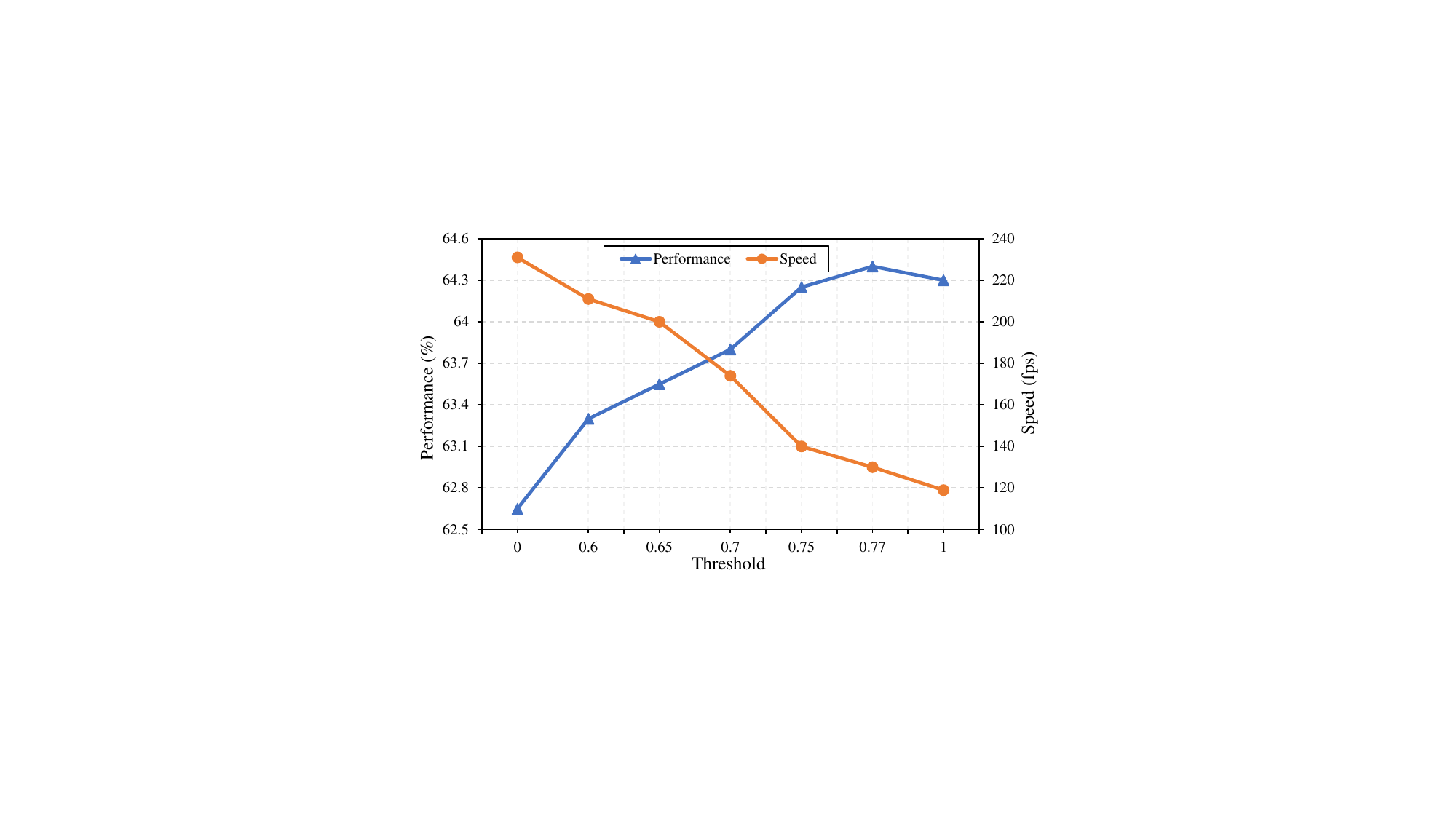}
\end{center}
   \caption{Impact of the scene complexity threshold on the speed and performance of DyHiT. Performance is measured as the average performance on LaSOT and GOT-10k, while speed is evaluated using a single-threaded test on the GOT-10K test set with an Nvidia GeForce RTX 2080 Ti GPU.}
\label{fig:threshold_influence}
\end{figure}

\noindent\textbf{Different Thresholds.} 
We use the scene complexity threshold, denoted as $\mathbf{T}$, to classify tracking scenarios. When the router predicts a score $\mathbf{F}$ greater than $\mathbf{T}$, the current scene is considered simple, and DyHiT utilizes Route1 for prediction. Otherwise, Route2 is employed. The value of $\mathbf{T}$ influences the speed and performance of DyHiT, as it determines the choice of route. As shown in Fig.~\ref{fig:threshold_influence}, adjusting $\mathbf{T}$ allows DyHiT to achieve a wide range of speed-accuracy trade-offs. When $\mathbf{T}$=0, only Route1 is used, resulting in a speed of 231 fps, which is $1.94 \times$ faster than HiT-Base, with an average performance of 62.6\% on LaSOT and GOT-10K. When $\mathbf{T}$=1, only Route2 is used, which is equivalent to HiT-Base. As the value of $\mathbf{T}$ increases, performance gradually improves, but speed decreases. When $\mathbf{T}$=0.77, DyHiT achieves its best performance of 64.4\%, slightly outperforming HiT-Base by 0.1\%, while maintaining a speed of 130 fps. The value of $\mathbf{T}$ can be adjusted according to specific use case requirements, allowing for a tailored speed-accuracy balance suited to the task at hand.

\begin{figure}[t]
\begin{center}
\includegraphics[width=1.0\linewidth]{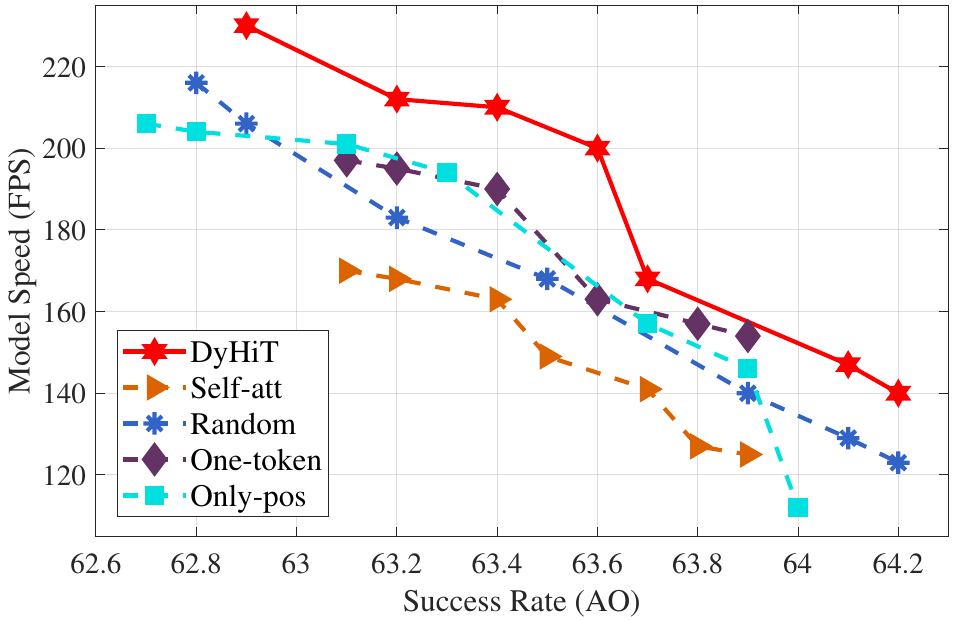}
\end{center}
   \caption{Comparison of different routers on GOT-10k in terms of speed (vertical axis) on Nvidia GeForce RTX 2080 Ti GPU and success rate (AO).}
\label{fig:different router}
\end{figure}

\begin{table}[t]
\centering
\caption{{
Comparison of different training settings on LaSOT (AUC). $\Delta$ denotes the performance (AUC) change (averaged over three trackers) compared with the baseline. We use \textcolor{gray}{gray} to denote baseline setting.}}
\label{ablation on training}
\setlength{\tabcolsep}{0.9mm}{
\begin{tabular}{l|l|cccc}
\toprule
\#&Method &DyOSTrack	&DyHiT$_{0.6}$	&DyHiT$_{0.65}$ &$\Delta$\\
\midrule[0.5pt]
\rowcolor{mygray1}
1&Baseline &69.5  &63.4 &63.5 &-\\
2& Wo-fixed, Separate   &68.8 &57.2 &62.7 &\textbf{-2.5} \\
3&Fixed, Joint &68.3 &56.6 &60.4 &\textbf{-3.7} \\
4&Reuse &67.9 &- &- &\textbf{-1.6}\\
\bottomrule
\end{tabular}}
\end{table}

\begin{figure*}[t]
\begin{center}
\includegraphics[width=1.0\linewidth]{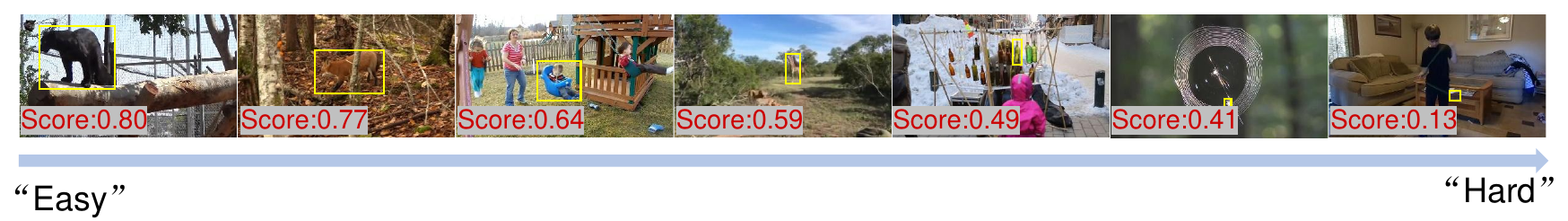}
\end{center}
   \caption{{Visualization of the Router Score. Zoom in for better visibility. }}
\label{fig:score_vis}
\end{figure*}

\begin{figure}[t]
\begin{center}
\includegraphics[width=1\linewidth]{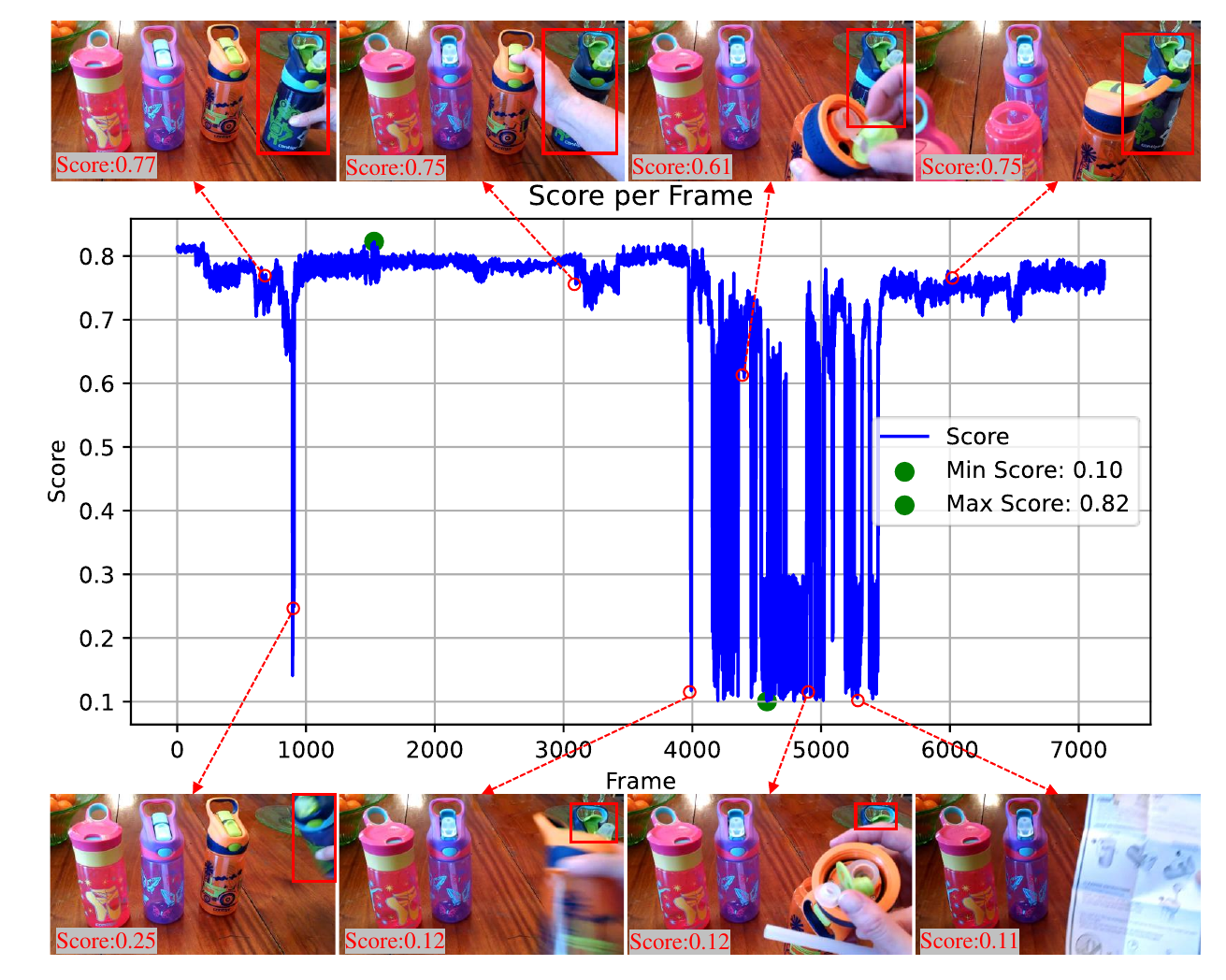}
\end{center}
   \caption{{Visualization of router scores across consecutive frames in a video sequence. The sequence is bottle-14 from the LaSOT dataset. Zoom in for better visibility. }}
\label{fig:video_score}
\end{figure}

\noindent\textbf{Different Routers.} To ensure the efficiency of the model, it is crucial to spend as little time as possible on determining the current scene, meaning our router should be concise and effective. As shown in Fig.~\ref{fig:different router}, we explore a total of five different routers, including our default setting DyHiT, Self-att method using a self-attention module to replace the linear layer, Random method using a random exit, One-token method employing only one token instead of the 256 tokens corresponding to the search region in our default setting, and Only-pos method that trains the router using only positive samples within the ground-truth bounding box to calculate the loss.
{By adjusting the scene classification threshold (ranging from 0.6 to 0.8) during inference, we can obtain a series of trackers with different speed-accuracy trade-offs.}
It can be observed that DyHiT achieves the best speed-accuracy trade-off. When obtaining the same Success Rate, The speed of DyHiT is faster than other methods. Taking a 63.2\% Success Rate as an example, The speed of DyHiT can reach 212 $fps$, which is 29 $fps$ faster than Random (183 $fps$), $23$ fps faster than One-token (189 $fps$), and 44 $fps$ faster than Self-att (168 $fps$). This demonstrates the simplicity and effectiveness of our default router.  Fig.~\ref{fig:score_vis} depicts the visualization results of the scores output by the router of our default DyHiT from simple to challenging scenes.
{Fig.~\ref{fig:video_score} presents a visualization of the router scores produced by our default DyHiT across a continuous video sequence, demonstrating the stability of our routing mechanism.}

\noindent\textbf{Different training settings.} We explore different training approaches for DyHiT, as shown in Table~\ref{ablation on training}. \#1 represents our default training approach described in Section~\ref{implementation}, where DyHiT undergoes a two-stage training with a frozen backbone.
\#2 indicates training without freezing the backbone in two stages, and \#3 denotes training with a frozen backbone but using a single stage to jointly train Route1 and the router in DyHiT. {\#4 represents the approach where we re-use the features from the efficient tracker to assist base tracker predictions in DyTracker and undergo training. Specifically, when DyTracker predicts in challenging scenes using the base tracker, we blend in features from the efficient tracker, originally used to measure scene difficulty. This is different from the default approach of directly using the results from the base tracker. 
}
We evaluate these different training approaches on three trackers, DyOSTrack, DyHiT$_{0.6}$, and DyHiT$_{0.65}$, on the LaSOT dataset. 
{In DyHiT, the subscript indicates the scene-splitting threshold: when the predicted score is higher than this threshold, the scene is classified as easy; otherwise, it is considered challenging.} 
Compared to \#2 and \#3, our default training approach shows an average improvement of 2.5\% and 3.7\% in AUC across the three trackers.
This demonstrates that the backbone trained in HiT is already capable of providing sufficiently accurate target features without the need for additional training. Additional training might lead to a performance decline. It also emphasizes the importance of training Route1 and the router in two stages. 
Combining the training of Route1 and the router in a single stage poses challenges, as inaccuracies in Route1 predictions could adversely affect router training. This can result in diminished discriminative capabilities of the router, leading to inaccurate assessments of scene difficulty. {Compared to the method in \#4, where features from the efficient tracker are re-used and trained, our approach of directly combining base tracker and DyHiT without additional training outperforms by 1.6\% in AUC. This suggests that the features extracted by the efficient tracker for challenging scenes might not be accurate, introducing noise when incorporated into the base tracker and leading to a performance decline. Therefore, we do not re-use the features output by the efficient tracker in complex scenes.
}

\begin{table}[h]
  \centering
      \caption{{Worst-case analysis on DyHiT. The reported speed refers to the inference speed of the tracker, excluding data pre-processing. }}
  \label{worst-cae}
  \setlength{\tabcolsep}{5mm}{
    \begin{tabular}{l|ccc}
    \toprule
    Method&GPU&CPU&AGX\\
    \midrule[0.5pt]
\cellcolor{mygray1}HiT-Base&\cellcolor{mygray1}175&\cellcolor{mygray1}33 &\cellcolor{mygray1}61\\
    \cellcolor{mygray2}DyHiT$_{1}$&\cellcolor{mygray2}171 &\cellcolor{mygray2}32 &\cellcolor{mygray2}56 \\
    \cellcolor{brown10}OSTrack~\cite{ostrack}&\cellcolor{brown10}105 &\cellcolor{brown10}11 &\cellcolor{brown10}19\\
    \cellcolor{brown20}DyOSTrack$_{1}$&\cellcolor{brown20}96 &\cellcolor{brown20}9 &\cellcolor{brown20}18 \\
    \cellcolor{mygray1}SeqTrack~\cite{SeqTrack}&\cellcolor{mygray1}31 &\cellcolor{mygray1}2 &\cellcolor{mygray1}6\\
    \cellcolor{mygray2}DySeqTrack$_{1}$&\cellcolor{mygray2}29 &\cellcolor{mygray2}2 &\cellcolor{mygray2}5 \\
    \bottomrule
    \end{tabular}
  }
\end{table}

\noindent{\textbf{Worst-case analysis.} The acceleration achieved by DyHiT primarily stems from leveraging a lightweight model that utilizes shallow features for prediction in simple scenarios. In contrast, for complex or challenging scenarios, the full model pipeline is activated to ensure robust performance. This dynamic routing mechanism, while effective, introduces additional computational overhead during the process of scene classification, which may lead to a reduction in inference speed under worst-case conditions.
To evaluate the performance impact under such conditions, we conducted worst-case analysis experiments, as reported in Table~\ref{worst-cae}. Specifically, DyHiT$_{1}$, DyOSTrack$_{1}$, and DySeqTrack$_{1}$ denote configurations where the full pipeline is used for evaluation, simulating the worst-case scenario. The results demonstrate that our router is highly efficient and introduces minimal overhead. For example, DyHiT$_{1}$ experiences only a minor drop of 4 $fps$, 1 $fps$, and 5 $fps$ compared to HiT-Base on GPU, CPU, and AGX respectively. Similarly, DySeqTrack$_{1}$ exhibits only a 2 $fps$ and 1 $fps$ decrease relative to SeqTrack on GPU and AGX, respectively.}

\begin{figure}[t]
\begin{center}
\includegraphics[width=1.0\linewidth]{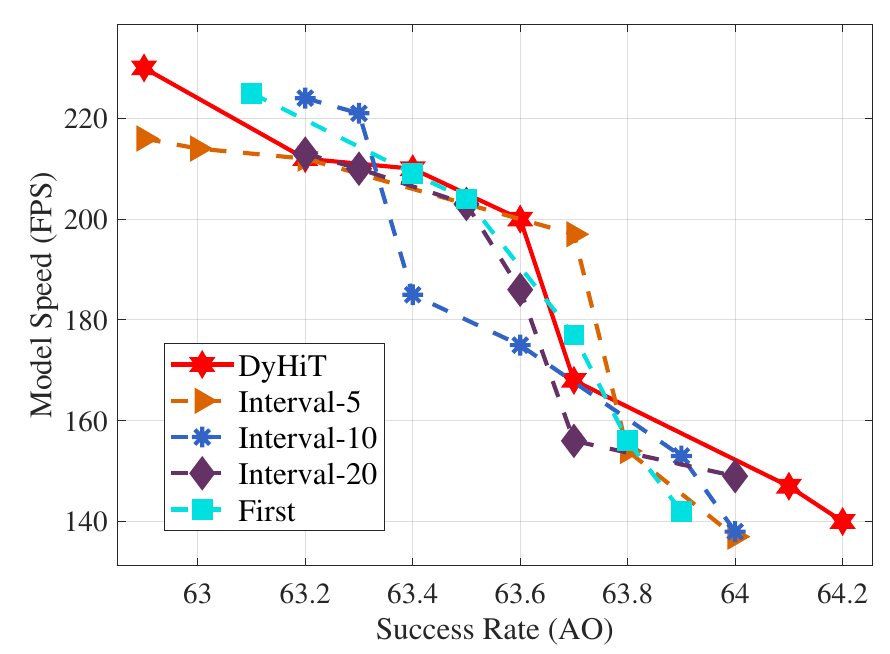}
\end{center}
   \caption{{Comparison of different scene classification frequencies on GOT-10k with respect to speed (vertical axis, measured on Nvidia GeForce RTX 2080 Ti GPU) and success rate (AO).}}
\label{fig:frequency}
\end{figure}
\noindent{\textbf{Different scene classification frequencies.} To investigate the impact of different scene classification frequencies on model performance, we conducted a series of experiments. The results are shown in Fig.~\ref{fig:frequency}. DyHiT refers to our default model, where scene classification is performed on every frame. Interval-5, Interval-10, and Interval-20 denote that scene classification is executed every 5, 10, and 20 frames, respectively. First indicates that classification is performed only once at the beginning of the video.
By adjusting the scene classification threshold  (ranging from 0.6 to 0.8) during inference, we can obtain a series of trackers with different speed-accuracy trade-offs.
As illustrated, DyHiT achieves the best trade-off between speed and accuracy. It reaches a maximum speed of 212 $fps$, and achieves the highest accuracy of 64.2\% AO score while maintaining a speed of 140 $fps$.}

\begin{table}[t]
\centering
\caption{{Statistics of scene occurrence frequency and corresponding speed on GOT-10k and LaSOT datasets. The speed is evaluated using a single-threaded test on the GOT-10K test set with an Nvidia GeForce RTX 2080 Ti GPU.}}
\label{scene frequency}
\setlength{\tabcolsep}{0.2mm}{
\begin{tabular}{l|ccccccc}
\toprule
 &DyHiT$_{0}$	&DyHiT$_{0.6}$	 &DyHiT$_{0.7}$ &DyHiT$_{1}$ &DyOSTrack\\
\midrule[0.5pt]
Easy-GOT&1 &0.83   &0.49  &0 &0.20\\
Hard-GOT&0 &0.17   &0.51  &1 &0.80\\
\midrule[0.5pt]
Easy-LaSOT&1 &0.82  &0.49  &0 &0.21\\
Hard-LaSOT&0 &0.18  &0.51  &1 &0.79\\
\midrule[0.5pt]
Speed/$fps$&231 &212   &175  &119 &110\\
\bottomrule
\end{tabular}}
\end{table}

\noindent{\textbf{Scene occurrence frequency.}
To more clearly demonstrate the primary source of DyHiT's speed improvement, we analyze the occurrence frequency of the two scene types on GOT-10k and LaSOT. The statistics are shown in Table~\ref{scene frequency}. Specifically, Easy-GOT and Hard-GOT represent the frequencies of easy and hard scenes on the GOT-10k dataset, respectively, while Easy-LaSOT and Hard-LaSOT denote the corresponding frequencies on the LaSOT dataset.
In DyHiT, the subscript indicates the scene-splitting threshold: when the predicted score is higher than this threshold, the scene is classified as easy; otherwise, it is considered challenging. When the subscript is set to 1, only Route2(full pipeline) is used; when set to 0, only Route1 is used.
As the threshold increases, the proportion of easy scenes gradually decreases, while the proportion of hard scenes increases. At the same time, the model's speed also gradually decreases. Specifically, DyHiT$_{0}$(Route1) achieves the highest speed of 231 $fps$, while DyHiT$_{1}$(Route2) reaches a speed of 119 $fps$. 
A scene-splitting threshold of 0.75 is adopted in DyOSTrack, resulting in a proportion of 0.2 for easy scenes and 0.8 for challenging ones.
}

\begin{table}[t]
  \centering
      \caption{{Ablation study on dynamic template. $\Delta_A$ denotes the performance change (averaged over benchmarks) compared with the baseline. 2T denotes dual templates. The speed is evaluated using a single threaded test on the GOT-10K test set with an Nvidia GeForce RTX 2080 Ti GPU.}}
  \label{dynamic-template}
  \setlength{\tabcolsep}{0.5mm}{
    \begin{tabular}{l|ccccc}
    \toprule
    Method&LaSOT(AUC)&GOT-10k(AO)&Speed($fps$)&$\Delta_A$\\
    \midrule[0.5pt]
\cellcolor{mygray1}HiT-Base&\cellcolor{mygray1}64.6&\cellcolor{mygray1}64.0 &\cellcolor{mygray1}126 &\cellcolor{mygray1}- \\
    \cellcolor{mygray2}HiT-Base-2T&\cellcolor{mygray2}64.8 &\cellcolor{mygray2}64.5 &\cellcolor{mygray2}117 &\cellcolor{mygray2}\textbf{+0.35} \\
    \cellcolor{brown10}DyHiT$_{0}$&\cellcolor{brown10}62.4 &\cellcolor{brown10}62.9 &\cellcolor{brown10}231 &\cellcolor{brown10}-\\
    \cellcolor{brown20}DyHiT$_{0}$-2T&\cellcolor{brown20}62.7 &\cellcolor{brown20}63.5 &\cellcolor{brown20}208 &\cellcolor{brown20}\textbf{+0.45}  \\
    \cellcolor{mygray1}DyHiT$_{0.65}$&\cellcolor{mygray1}63.5 &\cellcolor{mygray1}63.6 &\cellcolor{mygray1}200 &\cellcolor{mygray1}-\\
    \cellcolor{mygray2}DyHiT$_{0.65}$-2T&\cellcolor{mygray2}63.8 &\cellcolor{mygray2}64.0 &\cellcolor{mygray2}182 &\cellcolor{mygray2}\textbf{+0.35} \\
    \cellcolor{brown10}DyHiT$_{1}$&\cellcolor{brown10}64.6 &\cellcolor{brown10}64.0 &\cellcolor{brown10}119 &\cellcolor{brown10}-\\
    \cellcolor{brown20}DyHiT$_{1}$-2T&\cellcolor{brown20}64.8 &\cellcolor{brown20}64.5 &\cellcolor{brown20}110  &\cellcolor{brown20} \textbf{+0.35}\\
    \bottomrule
    \end{tabular}
  }
\end{table}

\noindent{\textbf{Dynamic Template.}
To address challenges such as target deformation and interference from similar objects in video sequences, we introduce a dynamic template mechanism based on the baseline model. We then analyze the impact of this dynamic template on both accuracy and speed. The results are presented in Table~\ref{dynamic-template}.
It can be observed that the incorporation of the dynamic template brings modest performance gains. The average performance on LaSOT and GOT-10k increases by 0.35\% for HiT-Base, DyHiT$_{0.65}$, and DyHiT$_{1}$, and by 0.45\% for DyHiT$_{0}$. However, these improvements come with a trade-off in speed due to the additional computational overhead introduced by the dynamic template, resulting in $fps$ drops of 9, 23, 18, and 9 for HiT-Base, DyHiT$_{0}$, DyHiT$_{0.65}$, and DyHiT$_{1}$, respectively.
}

\begin{figure}[t]
\begin{center}
\includegraphics[width=1.0\linewidth]{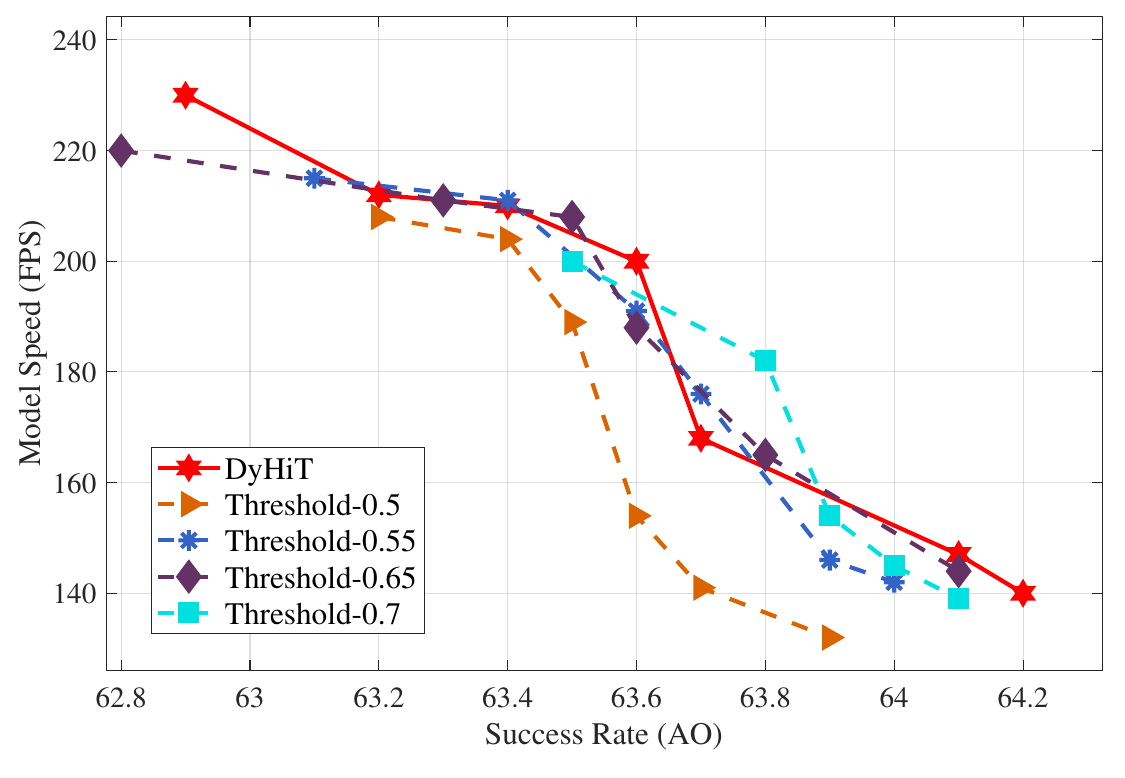}
\end{center}
   \caption{{Comparison of different foreground-background segmentation thresholds on GOT-10k with respect to speed (vertical axis, measured on Nvidia RTX 2080 Ti GPU) and success rate (AO).}}
\label{fig:score_threshold}
\end{figure}

\begin{figure}[t]
\begin{center}
\includegraphics[width=1.0\linewidth]{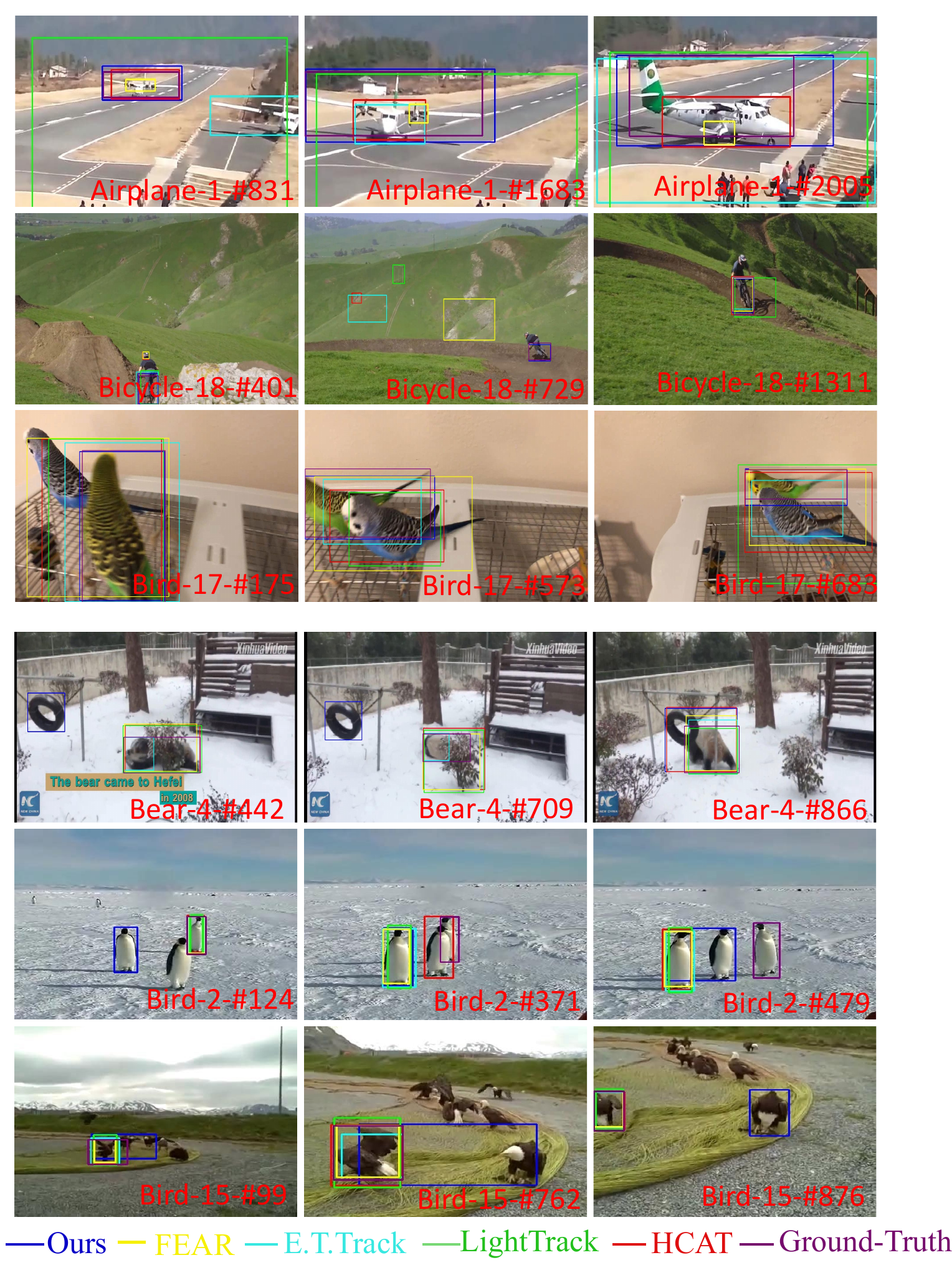}
\end{center}
   \caption{Qualitative comparison among HiT and other efficient trackers. The first three rows display successful tracking cases for HiT, while the following three rows show instances where tracking failed. The performance of HiT tends to degrade in the presence of distractions and cluttered backgrounds.}
\label{fig:vis_result}
\end{figure}

\noindent{\textbf{Different foreground-background segmentation thresholds.}
During the inference process, a threshold is employed to distinguish between foreground and background. We investigate the impact of five distinct thresholds on model performance. The results are presented in Fig.~\ref{fig:score_threshold}. DyHiT corresponds to the model using the default threshold of 0.6. Threshold-0.5, Threshold-0.55, Threshold-0.65, and Threshold-0.7 represent the utilization of 0.5, 0.55, 0.65, and 0.7 as the foreground-background segmentation thresholds, respectively. It is evident that DyHiT achieves a more extensive accuracy-speed trade-off. It reaches a maximum speed of 212 $fps$, and achieves the highest accuracy of 64.2\% AO score while maintaining a speed of 140 $fps$.
}

\section{Conclusion}
This study introduces HiT, a new family of efficient transformer-based tracking models. HiT addresses the disparity between tracking frameworks and lightweight hierarchical transformers through the proposed Bridge Module and dual-image position encoding. Building upon HiT, we further present DyHiT, a tracker capable of achieving a versatile range of speed-accuracy trade-offs using an efficient feature-driven dynamic routing architecture. 
Furthermore, we propose a training-free method based on DyHiT, to accelerate numerous high-performance trackers without compromising accuracy. Our extensive experiments demonstrate that our methods deliver promising performance within high speeds. We hope that this work could enhance the practical applicability of visual tracking and give insights for efficient visual tracking.

\noindent\textbf{Limitation.} One limitation of HiT is its challenge in handling distractors and background clutter, as illustrated in Fig.~\ref{fig:vis_result}. Additionally, our research primarily concentrates on closing the gap between lightweight hierarchical transformers and tracking frameworks. As a result, we make minimal adjustments to the existing hierarchical transformer without designing a new transformer specifically tailored for tracking.\\

\noindent\textbf{Acknowledgements.} The paper is supported in part by National Natural Science Foundations of China (No. U23A20384, No. 62176041, and No. 62402084), in part by Liaoning Province Science and Technology Plan (No. 2024JH2/102600040), and in part by China Postdoctoral Science Foundation (No. 2024M750319).

{
\small
\bibliographystyle{plain}
\bibliography{sn-bibliography}
}

\end{document}